\documentclass[sigconf]{acmart}

\AtBeginDocument{%
  }

\usepackage{pifont}%

\newcommand{\cmark}{\ding{51}}%
\newcommand{\xmark}{\ding{55}}%

\usepackage{caption}
\copyrightyear{2023}
\acmYear{2023}
\setcopyright{rightsretained}
\acmConference[MM '23]{Proceedings of the 31st ACM International Conference
on Multimedia}{October 29-November 3, 2023}{Ottawa, ON, Canada}

\acmBooktitle{Proceedings of the 31st ACM International Conference on
Multimedia (MM '23), October 29-November 3, 2023, Ottawa, ON, Canada}
\acmDOI{10.1145/3581783.3611737}
\acmISBN{979-8-4007-0108-5/23/10}

\usepackage{multirow}
\usepackage{tabularx}
\usepackage{colortbl}
\usepackage{booktabs} 
\usepackage{makecell}
\usepackage{graphbox}
\usepackage{footmisc}
\usepackage{bm}
\usepackage{caption}
\usepackage{arydshln}
\usepackage{dashrule}
\usepackage[labelformat=simple]{subcaption}
\usepackage{subcaption}
\usepackage{enumitem}
\setlist[itemize]{noitemsep,nolistsep}

\usepackage[capitalize]{cleveref}
\crefname{section}{Sec.}{Secs.}
\Crefname{section}{Section}{Sections}
\Crefname{table}{Table}{Tables}
\crefname{table}{Tab.}{Tabs.}
\Crefname{figure}{Figure}{Figures}
\crefname{figure}{Fig.}{Figs.}
\Crefname{equation}{Equation}{Equations}
\crefname{equation}{Eq.}{Eqs.}

\newcommand{\blue}[1]{\textbf{\textcolor{mblue}{#1}}}
\newcommand{\rblue}[1]{{\textcolor{mblue}{#1}}}

\newcommand{\nonecolor}[1]{}
\newcommand{\orange}[1]{\textbf{\textcolor{orange}{#1}}}

\newcommand{\bred}[1]{\textbf{\textcolor{red}{#1}}}

\newcommand{\bgreen}[1]{\textbf{\textcolor{mgreen}{#1}}}

\newcommand{\gray}[1]{\textcolor{mgray}{#1}}

\colorlet{lightcyan}{cyan!8}
\colorlet{lightpink}{pink!20}
\colorlet{lightgray}{gray!10}

\definecolor{mgray}{gray}{0.35}
\definecolor{mlgray}{gray}{0.85}

\definecolor{mred}{RGB}{238, 34, 12}
\definecolor{mgreen}{RGB}{1, 93, 0}
\definecolor{mblue}{RGB}{0, 77, 128}
\definecolor{mgreenblue}{RGB}{1,102,98}
\definecolor{orange}{RGB}{240, 120,0}

\begin{document}

\title{Towards Explainable In-the-Wild Video Quality Assessment:\\A Database and a Language-Prompted Approach}

\renewcommand{\shortauthors}{Haoning Wu et al.}

\begin{abstract}
The proliferation of in-the-wild videos has greatly expanded the Video Quality Assessment (VQA) problem. Unlike early definitions that usually focus on limited distortion types, VQA on in-the-wild videos is especially challenging as it could be affected by complicated factors, including various distortions and diverse contents. Though subjective studies have collected overall quality scores for these videos, how the abstract quality scores relate with specific factors is still obscure, hindering VQA methods from more concrete quality evaluations (\textit{e.g. sharpness of a video}).
To solve this problem, we collect over two million opinions on 4,543 in-the-wild videos on 13 dimensions of quality-related factors, including in-capture authentic distortions (\textit{e.g. motion blur, noise, flicker}), errors introduced by compression and transmission, and higher-level experiences on semantic contents and aesthetic issues (e.g. \textit{composition, camera trajectory}), to establish the multi-dimensional \textbf{Maxwell} database. Specifically, we ask the subjects to label among a positive, a negative, and a neutral choice for each dimension. These explanation-level opinions allow us to measure the relationships between specific quality factors and abstract subjective quality ratings, and to benchmark different categories of VQA algorithms on each dimension, so as to more comprehensively analyze their strengths and weaknesses. Furthermore, we propose the \textbf{MaxVQA}, a language-prompted VQA approach that modifies vision-language foundation model CLIP to better capture important quality issues as observed in our analyses. The MaxVQA can jointly evaluate various specific quality factors and final quality scores with state-of-the-art accuracy on all dimensions, and superb generalization ability on existing datasets. Code and data available at \url{https://github.com/VQAssessment/MaxVQA}.
\end{abstract}



\author{Haoning Wu}
\authornote{Both authors contributed equally to this research.}
\author{Erli Zhang}
\authornotemark[1]
\email{[haoning001, ezhang005]@e.ntu.edu.sg}
\affiliation{%
  \institution{\small{S-Lab, Nanyang Technological University}}
  \streetaddress{}
  \city{}
  \state{}
  \country{}
  \postcode{}
}

\author{Liang Liao}
\author{Chaofeng Chen}
\email{[liang.liao, chaofeng.chen]@ntu.edu.sg}
\affiliation{%
  \institution{\small{S-Lab, Nanyang Technological University}}
  \streetaddress{}
  \city{}
  \state{}
  \country{}
  \postcode{}
}

\author{Jingwen Hou}
\author{Annan Wang}
\email{[jingwen003, c190190]@e.ntu.edu.sg}
\affiliation{%
  \institution{\small{SCSE, Nanyang Technological University}}
  \streetaddress{}
  \city{}
  \state{}
  \country{}
  \postcode{}
}

\author{Wenxiu Sun}
\email{sunwx@tetras.ai}
\affiliation{%
  \institution{\small{SenseTime Group Limited}}
  \streetaddress{}
  \city{}
  \state{}
  \country{}
  \postcode{}
}

\author{Qiong Yan}
\email{yanqiong@tetras.ai}
\affiliation{%
  \institution{\small{Sensetime Research}}
  \streetaddress{}
  \city{}
  \state{}
  \country{}
  \postcode{}
}

\author{Weisi Lin}
\authornote{Corresponding Author.}
\email{wslin@e.ntu.edu.sg}
\affiliation{%
  \institution{\small{SCSE, Nanyang Technological University}}
  \streetaddress{}
  \city{}
  \state{}
  \country{}
  \postcode{}
}

\renewcommand{\shortauthors}{Haoning Wu et al.}


\received{4 May 2023}

\maketitle

\begin{figure}
    \centering
    \includegraphics[width=0.99\linewidth]{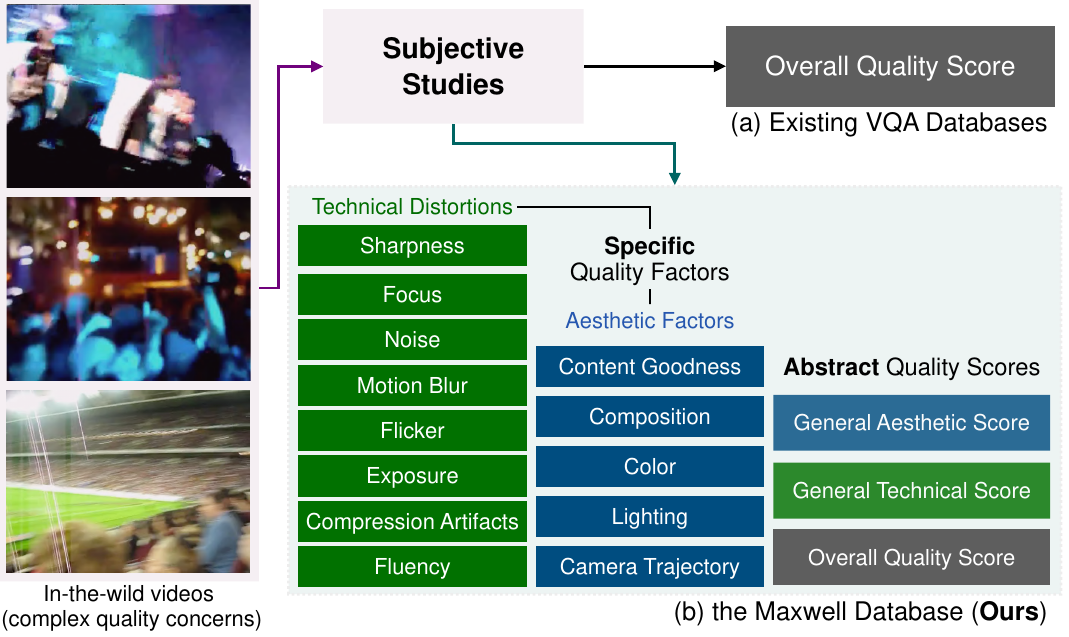}
    \vspace{-10pt}
    \caption{Quality for in-the-wild videos is influenced by complicated factors. For the first time, the proposed {Maxwell} database studies these \emph{specific} quality-related factors, and explore their impacts on the \emph{abstract} overall quality scores.}
    \label{fig:maxwell}
    \vspace{-15pt}
\end{figure}

\section{Introduction}

Rapid advances in technology have democratized the video production process, allowing more ordinary users to create and upload videos with smartphone cameras and public online platforms. Henceforth, Video Quality Assessment (VQA) on these prevalent in-the-wild videos is increasingly important for automatically recommending high-quality videos and improving low-quality ones.

Unlike traditional VQA~\cite{livevqa,csiqvqa} that focuses on specific types of distortions from pristine reference videos, VQA on in-the-wild videos is challenging as it is affected by authentic and commonly intermixed distortions; moreover, the contents of in-the-wild videos diverse and may also affect their quality. Although plenty of subjective studies~\cite{vqc,kv1k,pvq,cvd} have collected enormous quality scores on in-the-wild videos,  how specific quality-related factors (\textit{e.g. sharpness, exposure, fluency, noise}) affect or relate with the quality of a given video is still obscure. Consequently, existing objective VQA approaches~\cite{fastvqa,vsfa,tlvqm,videval} learnt from these scores are not able to evaluate these explainable concrete factors, thereby limiting their ability to provide targeted suggestions for video restoration or recommendation. Furthermore, the reliance on overall quality scores as the sole benchmark metric of VQA approaches makes it difficult to comprehensively analyze their ability on identifying specific quality issues, so as to apply them to appropriate scenarios.



To solve these challenges, we conduct a large-scale comprehensive subjective study to collect human opinions on specific quality factors, and explore how they affect the abstract quality scores. Specifically, we collect over two million annotations on 13 dimensions, each associated with a commonly-observed quality-related factor. These factors include in-capture authentic distortions~\cite{cvd,qualcomm} (\textit{e.g. blur, noise, poor exposure}), compression/transmission distortions~\cite{livevqa,csiqvqa}, and higher-level semantic-related (aesthetic) issues~\cite{vsfa,dover} (\textit{e.g. contents, color, composition}), as illustrated in Fig.~\ref{fig:maxwell}. To better collect subjective opinions on these dimensions, our study has several specific designs: \textit{First,} we notice that different factors can simultaneously affect the quality (such as Fig.~\ref{fig:maxwell}, which is fuzzy and with strong artifacts). To capture all factors that impact the quality of a given video, we prompt subjects to label all factors~\cite{spaq} for each video instead of choosing one major influencing factor~\cite{liqe,koniq}. \textit{Second,} in non-reference settings~\cite{rankiqa,metaiqa}, each specific factor (\textit{take sharpness as an example}) can pose either positive (\textit{while sharp}) or negative (\textit{while fuzzy}) impact to video quality, while it may also have no significant effects for some videos.
Thus, we define a forcefully positive-neutral-negative ternary choice for each specific factor, which ensures that subjects neither miss a factor for each video, nor face the dilemma of having to choose between good or bad when there are no significant effects. Moreover, to ensure the reliability of labels, we trained all participating subjects for each dimension before the annotation process and each axis in each video is annotated by 35 subjects following \cite{itu}. Correspondingly, we construct the \underline{M}ulti-\underline{Ax}is VQA Database \underline{w}ith \underline{E}xplanation-\underline{l}evel \underline{L}abels (\textbf{Maxwell}). The multi-dimensional opinions in the Maxwell database effectively bridge the gap between quality scores and specific quality factors, allowing for automatic and targeted repairment or enhancement on in-the-wild videos.

We further conduct an extensive analysis on the subjective opinions collected in the Maxwell database. Firstly, we find that the quality of in-the-wild videos is commonly affected by multiple factors, with an average of six factors affecting the quality of each video, which justifies our decision to label all factors. Secondly, we receive slightly more positive opinions than negative opinions, with a ratio of approximately 3:2. This suggests that it is appropriate to consider positive effects besides negative impacts for each quality factor. Thirdly, each factor affects the quality of at least 37\% of videos, justifying the validity of the selected specific factors.
Furthermore, our study revealed that human visual system is especially sensitive to temporal quality (Flicker and Fluency), emphasizing the importance of effective temporal modeling in VQA. The analysis on Maxwell helps us to better explain subjective VQA on in-the-wild videos, and further develop more effective objective approaches.


Inspired by the analysis, we propose the Multi-Axis Video Quality Assessor (\textbf{MaxVQA}), a language-prompted VQA approach that learns from comprehensive opinions in the Maxwell to jointly predict specific quality factors and overall quality, via integrating the vision-language foundation model CLIP~\cite{clip} with FAST-VQA~\cite{fastvqa}, the recent state-of-the-art VQA model for in-the-wild videos. Specifically, the MaxVQA first fuses CLIP visual features with FAST-VQA features to enhance the perception on low-level textures and temporal variations. It then compares the cosine similarity between the visual features and the textual features of a pair of dimension-specific prompts, so as to predict quality score for each dimension. Unlike existing strategies~\cite{spaq,koniq} that requires separate regressors for multi-objective quality evaluation, the MaxVQA unifies multiple dimensions with different text prompts. The proposed MaxVQA proves state-of-the-art accuracy not only on all dimensions in Maxwell, but also on overall quality assessment for existing VQA datasets. MaxVQA trained on Maxwell also shows well generalization ability on existing VQA datasets, suggesting its superb robustness. 

Our contributions can be summarized as three-fold: 

\begin{enumerate}[itemsep=2pt,topsep=0pt,parsep=0pt]
    \item We construct the \textbf{Maxwell} database, a comprehensive subjective study to collect over two million human opinions for 13 distinct specific quality factors on 4,543 in-the-wild videos. Our database allows for objective VQA methods to provide specific quality evaluations (\textit{e.g.} fluency of videos).
    \item Based on Maxwell, we analyze the characteristics of different factors, and their relations with overall quality scores for in-the-wild videos. Moreover, we provide the first multi-dimensional benchmark for objective VQA approaches to analyze their ability on capturing specific quality concerns.
    \item We design the \textbf{MaxVQA}, a vision-language-based VQA model that can jointly predict specific quality factors and final quality scores, by enhancing CLIP with low-level-sensitive FAST-VQA features. MaxVQA proves state-of-the-art performance and excellent generalization ability.
\end{enumerate}

\section{Related Work}

\subsection{Subjective Studies for In-the-Wild VQA}

 Unlike traditional settings~\cite{livevqa,csiqvqa} that mainly focus on compression~\cite{mdvqa,msu} or transmission-related distortions, subjective studies for in-the-wild VQA \cite{pvq,vqc,kv1k,ytugccc} directly collect human quality opinions on real-world videos. Thus, they face more complex quality-related issues~\cite{cvd,qualcomm,stablevqa,lightvqa} and much extended content diversity~\cite{rfugc,internetvqa,agiqa}. Therefore, these quality scores collected on in-the-wild videos can be affected by complicated reasons, and it is difficult to ground the effect of a given specific factor (\textit{e.g. sharpness or fluency of a video}). To investigate the relations between overall quality perception on in-the-wild videos and specific quality-related factors, we construct the \textbf{Maxwell} database that collects large-scale opinions for a wide range of specific quality factors together (\textit{distortions, content-related}) on in-the-wild videos, and associated analysis that better explains subjective VQA for in-the-wild videos. 

\subsection{Objective VQA Methods}

Recent years have witnessed rapid progresses of objective VQA methods~\cite{tlvqm,videval,fastvqa,vsfa,mdtvsfa,gstvqa,internetvqa,rirnet,bvqa2021,svqa}, which can be roughly categorized into two types: \textit{First}, classical methods, represented by TLVQM~\cite{tlvqm} and VIDEVAL~\cite{videval}, which design handcraft kernels to extract the low-level patterns of various common distortions that happen on in-the-wild videos. \textit{Second}, deep-learning-based methods. Pioneered by VSFA~\cite{vsfa}, these methods typically extract features from pre-trained deep neural networks, and regress them with overall quality opinions, such as recent BVQA-\textit{Li}~\cite{bvqa2021}. Most recently, FAST-VQA~\cite{fastvqa}, is proposed to learn quality scores end-to-end from pre-processed videos, and reaches superior accuracy with high efficiency. Although achieving higher accuracy, existing objective methods especially deep methods can hardly reason why their performance boost, nor predict the mechanism behind the quality scores. In recent years, vision-language foundation models~\cite{clip,align,xclip,coca,simvlm}, such as CLIP~\cite{clip} and ALIGN~\cite{align} can measure similarity between text sentences and images, while several studies~\cite{vila,liqe,clipiqa,clipiaa,bvqiplus,clip3dqa} have attempted vision-language modeling in VQA. With the constructed \textbf{Maxwell}, we integrate CLIP with FAST-VQA, and propose the language-prompted \textbf{MaxVQA} that can jointly evaluate multiple specific quality factors and overall quality scores. The MaxVQA can analyze video quality in a more explainable way.



\section{The Maxwell Database}

In this section, we elaborate on the designs and processes of the subjective study for the Maxwell database (Sec.~\ref{sec:maxwellproc}). The subjective study is conducted \textbf{in-lab} with 35 participants~\cite{itu} annotating for 4,543 videos on 16 dimensions, including 13 specific quality factors and 3 abstract quality ratings, as listed in Table~\ref{tab:dimpropmts}. Afterwards, we conduct extensive analyses on the collected subjective opinions, as discussed in Sec.~\ref{sec:maxwellanaop}. The well-designed subjective study in Maxwell as well as the extensive analyses help us to better investigate the effects of different specific factors on VQA for in-the-wild videos.

\subsection{Design of the Subjective Study}
\label{sec:maxwellproc}

\subsubsection{Choice of Quality-related Factors} In general, we conduct the study from two perspectives, the distortion-related technical perspective, and the semantic-related aesthetic perspective~\cite{dover}. We collect the general quality opinions from the two perspectives and choose several specific factors from these perspectives, as follows.

1) \bgreen{Factors in distortion-related technical perspective.} We collect opinions on common specific distortions (\textit{e.g. flicker, compression artifacts}) that happen in real-world videos. In general, the dimensions related to these factors have clear standards, that stronger distortion relates to worse quality~\cite{koniq,paq2piq}. Specifically, based on the technical origin of the distortions, they can be further grouped into \textit{in-capture authentic distortions}~\cite{cvd,qualcomm}, and \textit{post-capture distortions from compression or transmission}~\cite{icme2021,msu,mdvqa}. Specifically, we study six common~\cite{spaq,qualcomm} in-capture authentic distortions\footnote{To distinguish, we denote all \textit{temporal}-related factors with brackets, \textit{e.g.} \textit{[T-5]} Flicker. Other spatial factors (\textit{i.e. can be detected within a frame}) are in parenthesis.}:

\begin{itemize}
    \item [(T-1)] \textbf{Low Sharpness}: The video does not have clear textures.
    \item [(T-2)] \textbf{Out of Focus}: The salient target in video (\textit{e.g. human in a portrait video}) is not in-focus and looks Gaussian-blurred.
    \item [(T-3)] \textbf{Noise}: Random pixel-wise brightness or color variation.
    \item [(T-4)] \textbf{Motion Blur}: Blurriness that happens during and is caused by motions of camera or subjects in the video.
    \item [\textit{[T-5]}] \textbf{Flicker}: Non-smooth variation between adjacent frames. 
    \item [(T-6)] \textbf{Poor Exposure}: Unrecognizable regions of frames due to extremely low (high) brightness. 
\end{itemize}
and two common errors induced by compression or transmission:
\begin{itemize}
    \item [(T-7)] \textbf{Compression artifacts}: Block-like or moire-like artifacts introduced by compression algorithms~\cite{jpeg,h264}.
    \item [\textit{[T-8]}] \textbf{Low Fluency}: Missing frames during a moving sequence.
\end{itemize}

With specific distortions identified, we can employ existing respective algorithms to restore the video. For instance, if we detect low sharpness in a video, we can apply super-resolution~\cite{basicvsr} to repair it; denoising~\cite{fastdvdnet}, deblurring~\cite{gopro} and stablization~\cite{3dstabilize} can restore the video when noises, motion blurs or flickers are detected; frame interpolation~\cite{rife} can alleviate low fluency caused due to missing frames. These factor-level specific quality evaluations can assist automated video quality enhancement systems in the future.


 \begin{table}[]
    \centering
    \setlength\tabcolsep{10pt}
    \renewcommand\arraystretch{1.18}
    \caption{The codes and respective positive/negative descriptions during subjective study for different dimensions in the Maxwell database. }
    \vspace{-9pt}
    \resizebox{0.95\linewidth}{!}{\begin{tabular}{l|ccc} \hline
         Code & Dimension Names & \textit{Positive Description} & \textit{Negative Description} \\ \hline
        \bgreen{T-1} &  Sharpness & \textit{Sharp} & \textit{Fuzzy} \\
        \bgreen{T-2} &  Focus & \textit{In-Focus} & \textit{Out-of-Focus} \\
        \bgreen{T-3} &  Noise & \textit{Noiseless} & \textit{Noisy} \\
        \bgreen{T-4} &  Motion Blur & \textit{Clear-Motion} & \textit{Blurry-Motion} \\
        \bgreen{[T-5]} &  Flicker & \textit{Stable} & \textit{Shaky} \\
        \bgreen{T-6} &  Exposure & \textit{Well-exposed} & \textit{Poorly-exposed}  \\
        \bgreen{T-7} &  Compression Artifacts & \textit{Original} & \textit{Compressed}  \\
        \bgreen{[T-8]} &  Fluency & \textit{Fluent} & \textit{Choppy} \\  \hdashline
        \bgreen{*T-all} & \textit{Technical Perspective} & *Not Degraded & *Severely Degraded \\ \hdashline
                \blue{A-1} &  Contents & \textit{Good} & \textit{Bad} \\
        \blue{A-2} &  Composition & \textit{Organized} & \textit{Chaotic} \\
        \blue{A-3} &  Color & \textit{Vibrant} & \textit{Faded} \\
        \blue{A-4} &  Lighting & \textit{Contrastive} & \textit{Gloomy} \\
        \blue{[A-5]} &  (Camera) Trajectory & \textit{Consistent} & \textit{Incoherent} \\ \hdashline
        \blue{*A-all} & \textit{Aesthetic Perspective} & *Good Aesthetics & *Bad Aesthetics \\ \hdashline
        \textbf{*O} & \textit{Overall Quality score} & *High Quality & *Low Quality \\
         \hline
    \end{tabular}}
    \label{tab:dimpropmts}
    \vspace{-13pt}
\end{table}

2) \blue{Factors in semantic-related aesthetic perspective.} Many studies have suggested that visual quality is affected by semantic-related factors beyond technical distortions~\cite{dbcnn,vsfa,koniq,dover}. Unlike technical distortions, these higher-level factors are relatively under-studied in existing VQA researches. To better choose these semantic-related factors, we ask the subjects to score their feeling on videos based on four dimensions commonly concerned by existing image aesthetic assessment (IAA) studies~\cite{cadb,avaiaa,objiaa,distilliaa,piaadataset}, with an additional dimension for \textit{[temporal]} aesthetics on videos:
\begin{itemize}
    \item [(A-1)] \textbf{Contents}: Are the contents in the video appealing?
    \item [(A-2)] \textbf{Composition}: Do the video has organized and balanced composition of objects and scenes?
    \item [(A-3)] \textbf{Color}: Does the video has vibrant, pleasant color?
    \item [(A-4)] \textbf{Lighting}: Does the video has contrastive lighting?
    \item [\textit{[A-5]}] \textbf{Trajectory}: Does the camera moves in a consistent [temporal] trajectory that aligns with the scene?
\end{itemize}
\subsubsection{Design of the Study} 
After introducing the factors to study, we discuss the concrete form for the subjective study as follows.

1) \textbf{Evaluate the Impact of Each Factor.} Many existing studies~\cite{fastvqa,tlvqm} have suggested that different quality issues can concurrently exist and affect the quality of an in-the-wild video. Therefore, to conduct the study more comprehensively, instead of classifying one main factor that affect quality most significantly~\cite{liqe,koniq}, we ask subjects to evaluate the impact of each factor on video quality.

2) \textbf{\textit{Good/neutral/bad}: a ternary choice.} Though each factor will affect quality of certain videos, it is unlikely for all 13 factors to jointly significantly impact the quality of any one video. Therefore, we allow for a neutral choice for each dimension denoting that the corresponding factor does not notably impact perceptual quality of the video. Moreover, considering that each specific factor can pose either positive or negative impact to video quality, we design a ternary choice question for each dimension, including the neutral choice and a pair of antonyms to describe the positive and negative choices. The annotation form in our study is exemplified as follows:
\begin{verbatim}
(T-1) Sharpness:      Fuzzy < ----neutral---- > Sharp
(Please choose)       [ ]           [ ]           [ ]
\end{verbatim}
Descriptions for positive and negative choices are listed in Table~\ref{tab:dimpropmts}.

\begin{figure}
    \centering
    \includegraphics[width=0.88\linewidth]{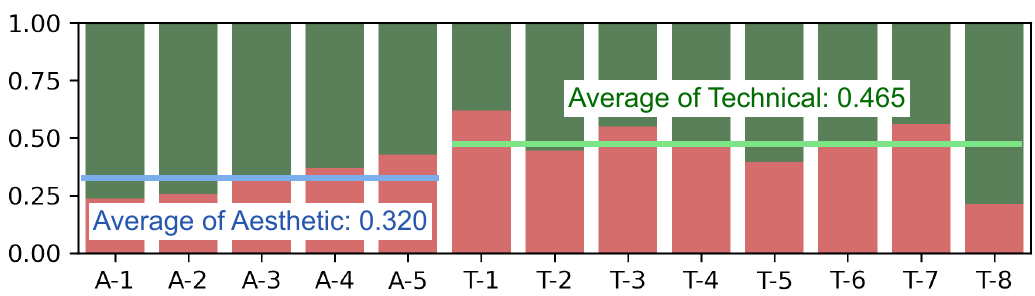}
    \vspace{-10pt}
    \caption{The tendency between negative (\textcolor{red}{red}) and positive (\textcolor{mgreen}{green}) opinions for each dimension. The overall average of positive-to-negative ratio is 1.44:1.}
    \label{fig:tendency}
    \vspace{-15pt}
\end{figure}

\subsubsection{Collection of Videos} To ensure the diversity of annotated videos, we collect them from large video databases~\cite{yfcc,k400data} and finally sub-sample 4,543 videos \cite{matchhistogram} that aligns with the distribution of the candidate set for annotation. The resolution of these videos ranges from 240P to 1080P, with an average duration 9s. 

\subsubsection{Training, Testing, and Annotation} To ensure that the subjects have clear and common understanding on all factor dimensions, we conduct the subjective studies \textbf{in-lab}. Moreover, before annotation, we collect three examples each for the positive (\textit{e.g. stable for Flicker axis}), negative (\textit{e.g. shaky for Flicker axis}), and neutral cases for all dimensions to \textbf{train} the participants. Moreover, similar as existing efforts~\cite{kv1k,pvq}, we also derive a testing process and reject the subjects that do not pass the testing. After rejection, every video has annotations from at least 31 accepted subjects. Denote the negative, neutral and positive opinions as $[-1, 0, 1]$, the mean factor opinion score ($\mathrm{MOS}_{a,i}$) for factor $a$ of video $i$ are obtained by averaging the raw opinions $\mathrm{OS}_{a,i}^k|_{k=0}^{K}$ from $K$ accepted subjects.


\subsection{Analyses on the Subjective Opinions}
\label{sec:maxwellanaop}

\begin{figure}
    \centering
    \includegraphics[width=0.9\linewidth]{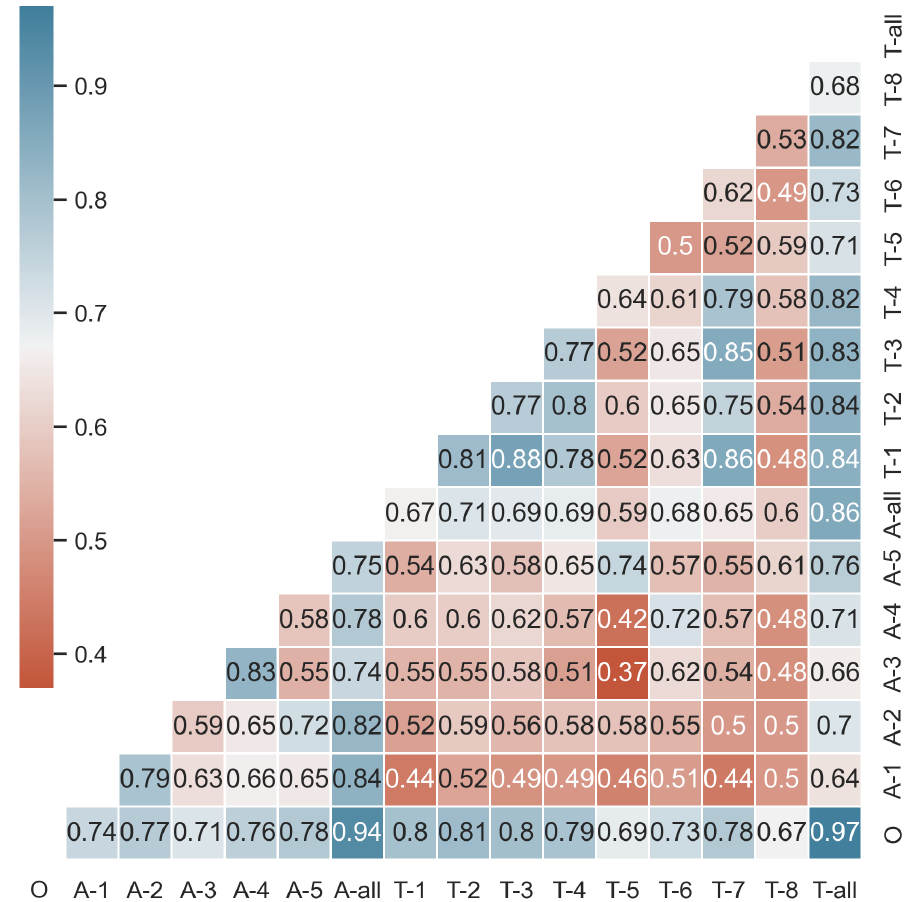}
    \vspace{-10pt}
    \caption{The correlation (\textit{Pearson Linear, PLCC}) map among various quality factors and overall quality scores. See Table~\ref{tab:dimpropmts} for full names for codes.}
    \label{fig:corr}
    \vspace{-15pt}
\end{figure}

\begin{figure*}
    \centering
 \includegraphics[width=0.89\linewidth]{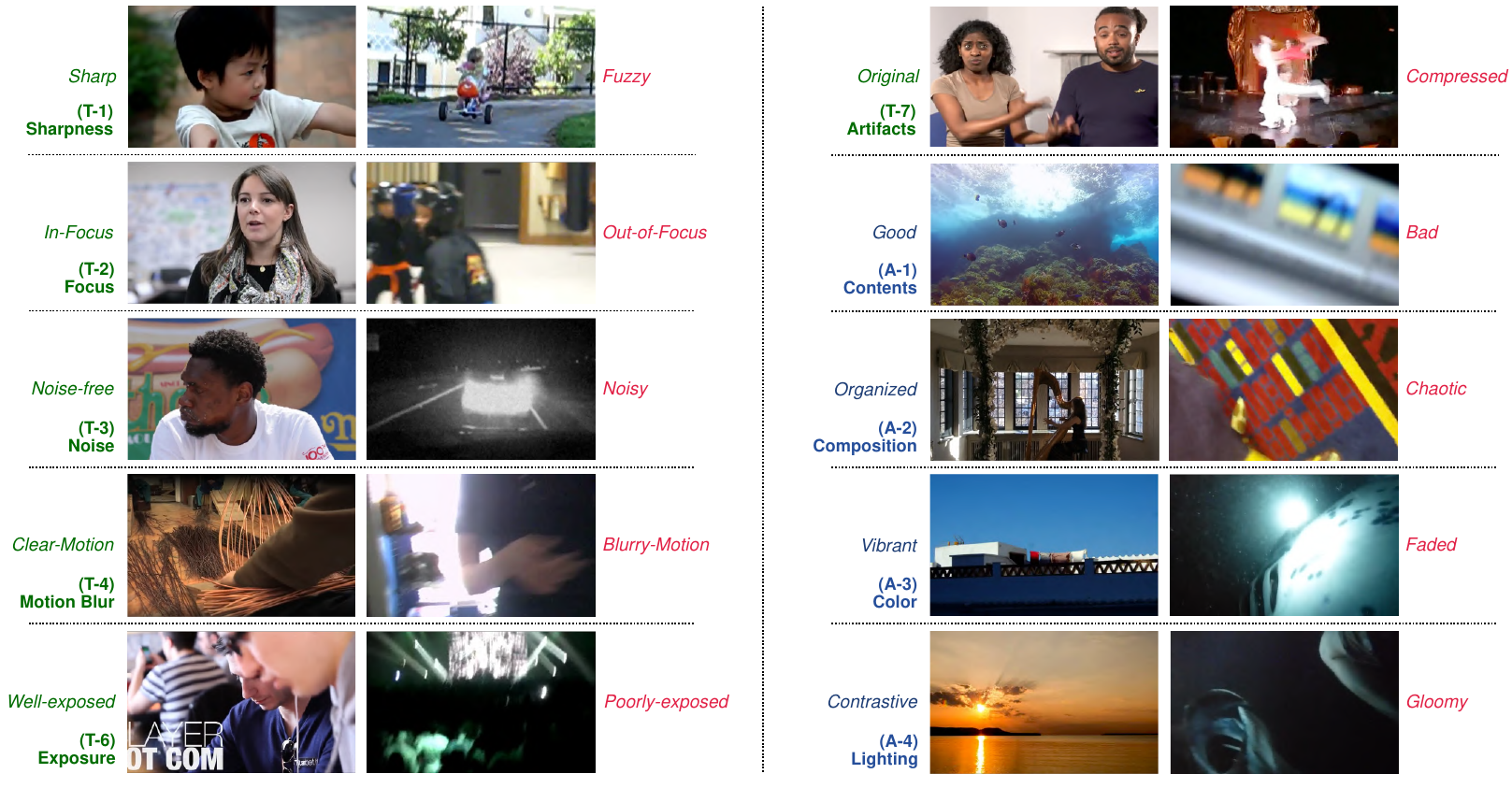}
    \vspace{-10pt}
    \caption{Qualitative studies on different specific factors, with a good video ($>$0.6) and a bad video ($<$-0.6) in each dimension of Maxwell; [A-5] Trajectory, [T-5] Flicker, and [T-8] Fluency are focusing on temporal variations and example videos for them are appended in supplementary package. Zoom in for details.}
    \label{fig:bestworst}
    \vspace{-13pt}
\end{figure*}

\begin{figure}
    \centering
    \includegraphics[width=0.88\linewidth]{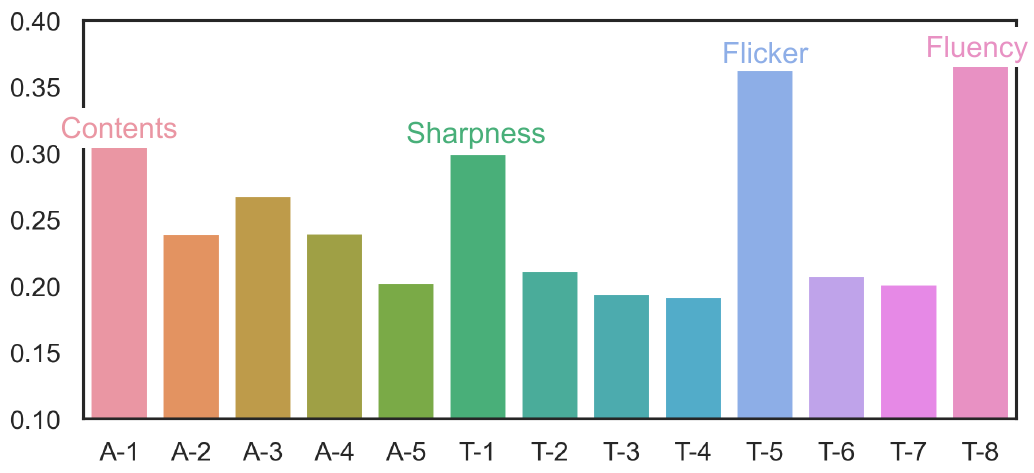}
    \vspace{-10pt}
    \caption{The absolute mean responses (AMR, \orange{co}\bgreen{lo}\blue{red}) and absolute raw responses (ARR, \gray{gray}) on diverse quality factors, where higher absolute response means subjects are more sensitive to the respective quality concerns.}
    \label{fig:arl}
    \vspace{-15pt}
\end{figure}

\subsubsection{Proportions of Different Opinions} First, we examine the proportion of non-neutral opinions in each dimension, which ranges from [0.37,0.56] with an average of 0.45. This means that the quality of an average video is affected by on-average \textbf{5.83} factors among the 13 factors, where each factor impacts the quality of \textbf{at least 37\%} of videos, proving the rationality of selecting these factors and labeling impact of each one. Moreover, we measure the tendency between positive and negative opinions through relative proportions after removing the neutral opinions, as shown in Fig.~\ref{fig:tendency}. The tendency for is balanced for technical distortions but significantly biased towards positive opinions for aesthetic factors. This result supports our design of the positive choices and suggests that human visual system may not only perceive \textit{quality} in a negative way.


\subsubsection{Correlation among Dimensions} Next, we visualize the correlation map of opinions among all dimensions (factors and abstract quality scores) in Fig.~\ref{fig:corr}, from which we notice several interesting observations. \textbf{\textit{First}}, all factors are highly relevant to, but also notably different from overall quality (\textbf{O}), with PLCC ranging in [0.67,0.81].  \textbf{\textit{Second}}, some distortions tend to happen together in real-world videos. For example, Sharpness (\bgreen{T-1}) most strongly associates with Noise (\bgreen{T-3}), which might be because low-definition video capturing devices may also be worse on dealing with noises. \textbf{\textit{Third}},
we observe a general low relevance between effects of aesthetic and technical factors (\textit{average PLCC about 0.5}). For example, the correlation between Color (\blue{A-3}) and Flicker (\bgreen{T-5})) is very low (0.37). This observation suggests that they usually have different influence to overall quality assessment on in-the-wild videos, supporting our design to divide these factors into two different perspectives during the subjective study.
\textbf{\textit{Fourth}}, the inter-relation between spatial and temporal distortions is also low, suggesting the importance of specific temporal modeling in VQA. These observations provide valuable guidance on improving future objective VQA models.


\subsubsection{Absolute Responses of Factors} In addition to correlation maps, we would also like to discover human sensitivity on different specific factors. Thus, we visualize the absolute mean responses ($\mathrm{AMR}_a$) in Fig.~\ref{fig:arl} as the database-wise average of \textbf{absolute} $\mathrm{MOS}_a$ for each factor $a$ in Maxwell, formulated as $\mathrm{AMR}_a = \frac{\sum_{i=1}^N |\mathrm{MOS}_{a,i}|}{N}$, where $\mathrm{MOS}_{a,i}$ is the mean factor opinion score of axis $a$ for video $i$, and $N=4,543$ is the number of videos in Maxwell. Furthermore, we visualize the proportion of non-neutral opinions in each dimension as absolute \textbf{raw} responses ($\mathrm{ARR}_a$) (\gray{gray} bars in Fig.~\ref{fig:arl}). From both responses, we observe the especial sensitivity on the two temporal distortions (Flicker (\bgreen{[T-5]}) and Fluency (\bgreen{[T-8]}), demonstrating that temporal modeling is important for real-world VQA. Moreover, Content (\blue{A-1}) and Sharpness (\bgreen{T-1}) are ranked next under both response metrics, suggesting that they are also important quality concerns to be considered for in-the-wild videos.


\subsubsection{Qualitative Studies on Specific Factors} In Fig.~\ref{fig:bestworst}, we further illustrate extreme examples in each dimension, to better qualitatively understand the quality concerns of different dimensions. From these examples, we validate that technical distortions are usually intermixed in real-world videos. Moreover, the semantic-related aesthetic dimensions are also highly associated with overall quality perception, such as Composition, Color, and Lighting (where the \textit{good} and \textit{bad} cases are with obvious higher and lower quality). In summary, different dimensions represent diverged quality concerns which align with conventional definitions about them, providing reliable supervisions for objective specific quality evaluation.

\begin{figure*}
    \centering
    \includegraphics[width=0.94\linewidth]{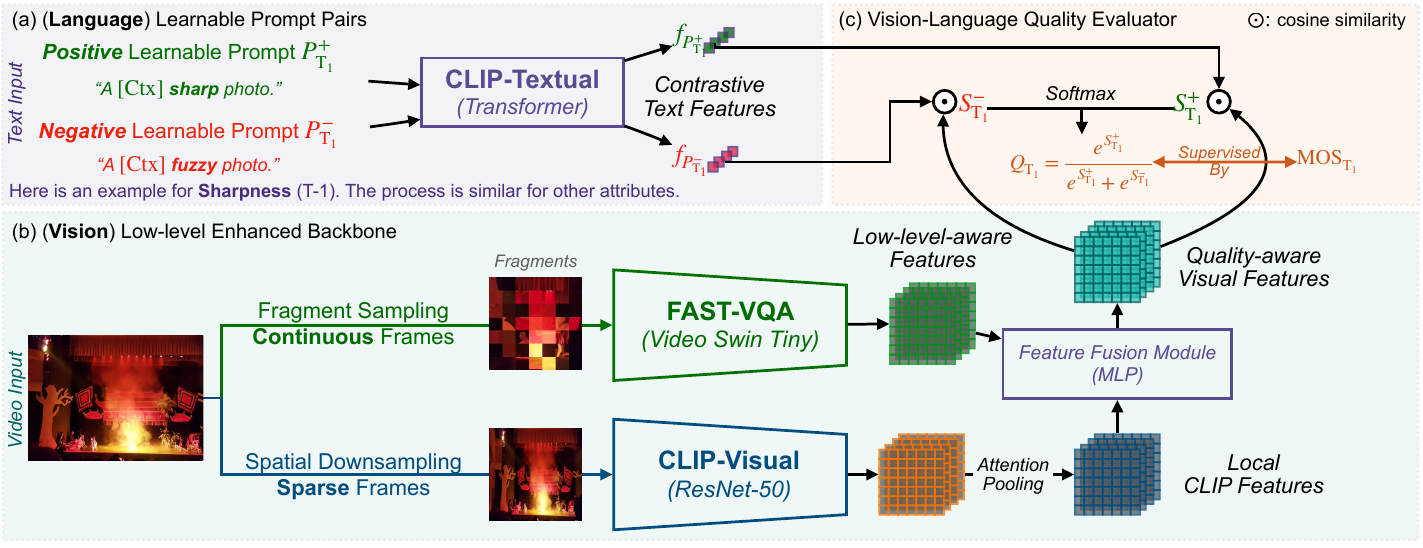}
    \vspace{-10pt}
    \caption{The structure of the proposed \underline{M}ulti-\underline{ax}is \underline{V}ideo \underline{Q}uality \underline{A}ssessor (\textbf{MaxVQA}), including (a) Learnable Contrastive Language Prompts to \textit{encode text inputs}, (b) Low-level Enhanced Visual Backbone to \textit{encode videos}, (c) and the final Vision-Language Quality Evaluator to \textit{output multi-axis quality scores}.}
    \label{fig:maxvqa}
    \vspace{-10pt}
\end{figure*}

\section{Our Approach: The MaxVQA}

\subsection{Overview: Language-Prompted VQA}


\subsubsection{Unifying Dimensions via Language Prompts} With large-scale multi-dimensional human opinions collected in the Maxwell database, we would like to design an objective approach that can learn from these opinions and then jointly predict specific quality factors and overall quality scores for in-the-wild videos. Unlike traditional multi-task strategies~\cite{spaq,koniq} that regress each training objective independently, we propose to unify these inter-related dimensions via a language-prompted paradigm. Specifically, it is based on CLIP (Contrastive Language Image Pre-training)~\cite{clip}, which is composed of a textual encoder (\textbf{CLIP-Textual}, $\mathrm{E}_t$) and a visual encoder (\textbf{CLIP-Visual}, $\mathrm{E}_v$). Given any text prompt $P$ and video $\mathcal{V}=\{V_t|_{t = 0}^{N-1}\}$ as inputs, CLIP can encode them into the same representation space:
\begin{equation}
    f_{V,t} = \mathrm{E}_v(V_t)|_{t = 0}^{N-1};~~~~~~~~f_{P} = \mathrm{E}_t(P)
\end{equation}
Then, we can calculate the similarity between prompt $P$ and $V_t$: 
\begin{equation}
    \mathrm{Sim}(P, V_t) = \frac{f_{V_t}\cdot f_{P}}{\Vert f_{V_t}\Vert\Vert f_{P}\Vert}
\end{equation}
which allows us to unify quality evaluation on different dimensions by setting different text prompts $P$. Details are discussed in Sec.~\ref{sec:vlqe}.

\subsubsection{Modifying CLIP for in-the-wild VQA} \label{sec:modifyclip} Though CLIP can jointly encode videos and various natural language text prompts, it still has several problems in both its visual and textual parts hindering it from modeling VQA more effectively. {\textbf{1) For the visual part}}, due to downsampling and global pooling operations, CLIP-visual has reduced sensitivity to low-level detail-related factors such as \textit{noise, sharpness, artifacts}, which are most correlated with overall quality scores. Moreover, CLIP-visual does not have any temporal modeling, which neglects the temporal distortions as proved important in our analysis. To solve the problems, we propose to utilize the local CLIP visual features by modifying its attention pooling outputs (Sec.~\ref{sec:map}); more importantly, we fuse the CLIP-visual features with detail-aware and temporal-aware FAST-VQA features (Sec.~\ref{sec:fastvqa}). {\textbf{2) For the textual part,}} previous attempts have shown that unlike general prompts (\textit{e.g. good/bad}) that can effectively match overall quality scores, the prompts for specific factors are poorly aligned with human perception (see Table~\ref{tab:ablation}). Therefore, we adopt the learnable contextual prompts to optimize the textual inputs (Sec.~\ref{sec:icp}). With these designs, we propose the \underline{M}ulti-\underline{ax}is \underline{V}ideo \underline{Q}uality \underline{A}ssessor (\textbf{MaxVQA}), discussed as follows.






\subsection{Low-level Enhanced Visual Backbone}

In this section, we introduce the enhanced visual backbone in MaxVQA (Fig.~\ref{fig:maxvqa}(b)), which extracts the local CLIP features after attention pooling, and further fuses them with FAST-VQA features to enhance CLIP-visual on low-level and temporal distortions.

\subsubsection{Local Features from CLIP}
\label{sec:map}

For the original CLIP, the visual backbone encodes the global embedding $f_{V_t}$ for each frame through the attention pooling layer. Denote the feature before the specific attention pooling layer in CLIP-ResNet-50 backbone as $f_{V_t}^\text{Pre-Pool}$, this final attention pooling is achieved by a single-layer multi-head self-attention ($\mathrm{MHSA}$) on these features, as follows:
\begin{equation}
    [f_{V_t}, f_{V_t}^\text{Local}] = \mathrm{MHSA} ( [\overline{f_{V_t}^\text{Pre-Pool}}, f_{V_t}^\text{Pre-Pool}])
\end{equation}
while only $f_{V_t}$ is used in naive CLIP, we notice that the output local features $f_{V_t}^\text{Local}$ could also be useful as most quality factors as studied in the Maxwell can be localized. Moreover, existing study~\cite{denseclip} proves that these local features are more sensitive to object-level recognition (\textit{detection, segmentation}), which are also associated with aesthetic-related dimensions~\cite{objiaa,distilliaa}. Thus, we take the $f_{V_t}^\text{Local}$ instead of $f_{V_t}$ as output features of CLIP-visual.

\begin{table*}
\footnotesize
\caption{Multi-Axis benchmark comparison of existing approaches and the proposed MaxVQA on Maxwell. [{Temporal}] dimensions are labeled in brackets.}\label{tab:existing}
\vspace{-4pt}\setlength\tabcolsep{6pt}
\renewcommand\arraystretch{1.19}
\footnotesize
\centering
\vspace{-8pt}
\resizebox{\linewidth}{!}{\begin{tabular}{l|c|c|c|c|c|c|c|c|c|c|c|c|c|c|c|c}
\hline
   Dimensions (\textit{in codes})  & \blue{A-1} & \blue{A-2} & \blue{A-3} & \blue{A-4} & \blue{[A-5]} & \blue{A-all} & \bgreen{T-1} & \bgreen{T-2} & \bgreen{T-3} & \bgreen{T-4} & \bgreen{[T-5]} & \bgreen{T-6} & \bgreen{T-7} & \bgreen{[T-8]} & \bgreen{T-all} & \textbf{O}    \\ \hline
  Methods           & PLCC     & PLCC   & PLCC   & PLCC   & PLCC   & PLCC   & PLCC   & PLCC   & PLCC   & PLCC   & PLCC   & PLCC   & PLCC   & PLCC   & PLCC   & PLCC  \\ \hline 


\multicolumn{17}{l}{\textbf{\textit{Zero-shot Methods:}} (not fine-tuned on any dimensions)} \\ \hdashline
\textit{(c, spatial)} NIQE~\cite{niqe} & 0.317 & 0.281 & 0.329 & 0.321 & 0.211 & 0.338 & 0.189 & 0.255 & 0.174 & 0.217 & 0.136 & 0.199 & 0.156 & 0.178 & 0.255 & 0.301 \\
\textit{(c, temporal)} TPQI~\cite{tpqi} & 0.246 & 0.293 & 0.210 & 0.225 & 0.360 & 0.319 & 0.223 & 0.293 & 0.239 & 0.374 & \textbf{0.463} & 0.225 & 0.244 & \textbf{0.410} & 0.363 & 0.361 \\
\textit{(CLIP-based)} SAQI~\cite{buonavista} &\textbf{0.388}&\textbf{0.410}&\textbf{0.453}&\textbf{0.504}&\textbf{0.393}&\textbf{0.515}&\textbf{0.560}&\textbf{0.500}&\textbf{0.524}&\textbf{0.509}&0.344&\textbf{0.482}&\textbf{0.497}&0.311&\textbf{0.554}&\textbf{0.559}\\
\hline
\multicolumn{17}{l}{\textbf{\textit{Supervised Methods:}} (for existing approaches, we adopt naive multi-task training on all dimensions)} \\ \hdashline
\textit{(classical)} TLVQM\cite{tlvqm}  &0.477&0.523&0.437&0.471&0.601&0.590&0.537&0.571&0.538&0.606&0.664&0.503&0.539&0.530&0.653&0.652  \\

\textit{(classical)} VIDEVAL\cite{videval} & 0.469 & 0.533 & 0.501 & 0.513 & 0.533 & 0.564 &  0.578 & 0.534 & 0.548 &0.557 & 0.664 &  0.467 & 0.543 & 0.393 & 0.595 & 0.601        \\ \hdashline
\textit{(c+d)} RAPIQUE\cite{rapique} &0.490&0.538&0.520&0.559&0.560&0.651&0.610&0.618&0.588&0.621&0.563&0.568&0.566&0.406&0.695&0.708  \\ 
\hdashline
\textit{(deep)} VSFA\cite{vsfa}   & 0.512 & 0.556 & 0.611 & 0.634 & 0.515 & 0.624 & 0.719 & 0.625 & 0.642 & 0.612 & 0.555 & 0.645 &  0.643 & 0.406 & 0.672 & 0.678 \\
\textit{(deep)} BVQA-\textit{Li}\cite{bvqa2021} &0.553&0.607&0.659&0.668&\underline{0.678}&0.671&0.746&0.686&0.694&0.682&\underline{0.781}&0.653&0.677&\underline{0.659}&0.759&0.739 \\ 
\textit{(deep)} {FAST-VQA}\cite{fastvqa} & \underline{0.614} & \underline{0.630} & \underline{0.696} & \underline{0.709} & 0.646 & \underline{0.721} & \underline{0.800} & \underline{0.724} & \underline{0.755} & \underline{0.731} & {0.751} & \underline{0.695} & \underline{0.736} & {0.654} & \underline{0.803}  &  \underline{0.782} \\ \hline
\textbf{MaxVQA} (Ours) &\textbf{0.681}&\textbf{0.701}&\textbf{0.757}&\textbf{0.749}&\textbf{0.712}&\textbf{0.775}&\textbf{0.825}&\textbf{0.748}&\textbf{0.776}&\textbf{0.761}&\textbf{0.782}&\textbf{0.748}&\textbf{0.763}&\textbf{0.684}&\textbf{0.827}&\textbf{0.813} \\
\hline
\end{tabular}}

\vspace{-6pt}
\end{table*}

\subsubsection{Fusion with FAST-VQA Features}
\label{sec:fastvqa}

Many existing works~\cite{videval,fastvqa,rapique,cnntlvqm} have pointed out that directly using fixed features from high-level pre-trained deep neural networks~\cite{he2016residual,vit} may have compromised sensitivity on several texture-related factors (\textit{e.g. noises, artifacts and sharpness}) due to downsampling of visual inputs. To enhance these important low-level perceptual factors and compliment the lack of temporal modeling in CLIP, we fuse the CLIP feature with the FAST-VQA~\cite{fastvqa,fastervqa} features, the state-of-the-art VQA-specific features which proved excellent performance on several VQA datasets and well distinction on temporal distortions. To get these features, the videos are passed through the fragment sampling (\textit{crop multiple \textbf{original resolution} patches and splice them together}, see in Fig.~\ref{fig:maxvqa}(b)), and then fed into a modified Video Swin Transformer~\cite{swin3d}. Denote the fragment sampling as $\mathbf{F}$, the FAST-VQA features $F_{\mathcal{V}}^\text{FAST}$ of $\mathcal{V}=\{V_t|_{t = 0}^{N-1}\}$ are extracted as follows:
\begin{equation}
    F_{\mathcal{V}}^\text{FAST} = \mathrm{Swin}([\{\mathbf{F}(V_t)|_{t = 0}^{N-1}\}])
\end{equation}
where $\mathrm{Swin}$ denotes the modified Video Swin Transformer Tiny, that takes temporally-aligned fragments (concatenated as videos) as inputs. \textit{More details for fragment sampling are in supplementary.}

Finally, the FAST-VQA features are fused with local CLIP-Visual features through a residual multi-layer perceptron ($\mathrm{MLP}$):
\begin{equation}
    f_{V_{t}}^\text{Final} = \mathrm{MLP}([f_{V_{t}}^\text{Local}, (f_{\mathcal{V}}^\text{FAST})_t])+ f_{V_{t}}^\text{Local}
\end{equation}
where $(f_{\mathcal{V}}^\text{FAST})_t$ denotes $t$-th feature frame of $f_{\mathcal{V}}^\text{FAST}$; $t\in [0,N)$.

\subsection{Learnable Language Prompts}
\label{sec:icp} 

To distinguish diverse dimensions, we initialize the text prompts differently for each axis with their respective positive and negative descriptions in Maxwell (see Table~\ref{tab:dimpropmts}). Denote positive and negative descriptions for $a$ as $D_a^{Pos}$ and $D_a^{Neg}$, the initial positive $\hat P_a^{+}$ and negative $\hat P_a^{-}$ prompts for a technical factor $a$ (T-x) are defined as:
\begin{equation}
    \hat P_a^{+} =  D_a^{Pos} +  \text{`` photo.''};~~~~~~~\hat P_a^{-} =  D_a^{Neg} + \text{`` photo.''}
\end{equation}

For the semantic-related aesthetic factors, the prompts are initialized with their dimension name (denoted as $\mathrm{Name}_a$, \textit{e.g.} Contents for A-1, Color for A-3) for more targeted and accurate modeling:
\begin{equation}
    \hat P_a^{+} =  D_a^{Pos} + \mathrm{Name}_a +  \text{`` photo.''};~~~~~~~\hat P_a^{-} =  D_a^{Neg} + \mathrm{Name}_a + \text{`` photo.''}
\end{equation}

Though some more abstract initial prompt pairs (\textit{e.g. good/bad}) have proved good performance~\cite{liqe,clipiqa,buonavista} on overall quality perception, both existing studies and our experiments (Table~\ref{tab:ablation}, \textit{row 1}) suggest that the more specific prompts are poorly aligned with respective human perception. Therefore, we choose the simple and efficient contextual prompt~\cite{coop} to optimize these initial prompts, by inserting a single context token before the initial prompts:
\begin{equation}
    P = \text{``A ''} + \mathrm{Ctx} + \hat P
\end{equation}
where $\mathrm{Ctx}$ is the context token, initialized as \textit{``X"} and optimized during training, and $\hat P$ is an overall denotion of $P_a^{+}$ and $P_a^{-}$. Tokens except $\mathrm{Ctx}$ and weights of the language encoder are keep \textbf{frozen}.

\begin{table}
\footnotesize
\caption{Evaluation of MaxVQA on existing in-the-wild VQA datasets. All experiments are conducted under 10 train-test splits with mean results reported.}\label{tab:existing}
\vspace{-4pt}\setlength\tabcolsep{4.5pt}
\renewcommand\arraystretch{1.25}
\footnotesize
\centering
\vspace{-8pt}
\resizebox{\linewidth}{!}{\begin{tabular}{l|cc|cc|cc}
\hline
   Dataset  & \multicolumn{2}{c|}{{LIVE-VQC}}   & \multicolumn{2}{c|}{{KoNViD-1k}}    & \multicolumn{2}{c}{{YouTube-UGC}}      \\ \hline
  Methods           & SRCC$\uparrow$& PLCC$\uparrow$  & SRCC$\uparrow$& PLCC$\uparrow$         & SRCC$\uparrow$& PLCC$\uparrow$       \\ \hline 

TLVQM\cite{tlvqm}  & 0.799 &  0.803  & 0.773 & 0.768   & 0.669 &  0.659    \\

VIDEVAL\cite{videval} & 0.752 &  0.751  & 0.783 & 0.780           & 0.779 &  0.773       \\ \hdashline
RAPIQUE\cite{rapique}    & 0.755 &  0.786  & 0.803 & 0.817                & 0.759 &  0.768     \\ 
CNN+TLVQM\cite{cnntlvqm}   & 0.825 & 0.834 & 0.816 & 0.818  & 0.809 & 0.802   \\
\hdashline
VSFA\cite{vsfa}    & 0.773 &  0.795  & 0.773 & 0.775   & 0.724 &  0.743  \\
PVQ\cite{pvq}   & {0.827} &  {0.837}  & 0.791 &   0.786        & NA &  NA           \\
CoINVQ\cite{rfugc}  & NA &  NA & 0.767 &  0.764    & {0.816} &     {0.802}   \\ 
BVQA-\textit{Li}\cite{bvqa2021} & 0.834 & 0.842 & 0.834 & 0.836 & 0.818 & 0.826  \\ 
DisCoVQA\cite{discovqa} &  0.820 & 0.826 & 0.846 & 0.849 &  0.809 & 0.808 \\ 
{FAST-VQA}\cite{fastvqa} &  {\underline{0.849}} & {\underline{0.862}} & {\underline{0.891}} & {\underline{0.892}} & {\underline{0.855}} & {\underline{0.852}} \\ \hline
\textbf{MaxVQA} (Ours) & \textbf{0.854} & \textbf{0.873}  & \textbf{0.894} &  \textbf{0.895}  & \textbf{0.894} & \textbf{0.890} \\
\hline
\end{tabular}}
\vspace{-8pt}
\end{table}

\subsection{Unified Vision-Language Quality Evaluator}
\label{sec:vlqe}

To finally evaluate quality, the proposed MaxVQA calculates the positive similarity ($S_a^+$) and negative similarity ($S_a^-$) in dimension $a$ from its specific positive $P_a^+$ and negative $P_a^-$ prompts,
\begin{equation}
    S_a^+ = \sum_{t=0}^{N-1} \frac{\mathrm{Sim}(P_a^+, f_{V_t}^\text{Final})}{N};~~~~~~~~~~S_a^- = \sum_{t=0}^{N-1} \frac{\mathrm{Sim}(P_a^-, f_{V_t}^\text{Final})}{N}
\end{equation}
and perform softmax pooling to obtain the respective quality scores:
\begin{equation}
    Q_{a}=\frac{e^{S_a^+}}{e^{S_a^+}+e^{S^-_a}}
\end{equation}

\section{Experimental Analysis}

In this section, we benchmark existing VQA approaches on the Maxwell database (Sec.~\ref{sec:bench}), and evaluate the performance and generalization ability of the proposed MaxVQA (Sec.~\ref{sec:bench}\&\ref{sec:existing}). We also conduct ablation studies (Sec.~\ref{sec:abl}) and qualitative studies (Sec.~\ref{sec:qual}) to further analyze the effectiveness of the proposed MaxVQA.

\subsection{Implementation Details} 

\subsubsection{Experimental Setups} We evaluate the proposed MaxVQA with \textbf{frozen} visual and textual encoders, and pre-extract the visual features to reduce computational cost. The only optimizable parameters are the MLP module for visual feature fusion and the contextual prompt $\mathrm{Ctx}$, therefore the training requires only 3GB graphic memory cost at batch size 16. Following original CLIP, the videos are downsampled to $224\times224$ before fed to $\mathrm{E}_v$. Our code is based on OpenCLIP~\cite{openclip}. The CLIP-visual backbone is ResNet-50.

\subsubsection{Database Settings} Following recent practices~\cite{msu,pvq}, we split the Maxwell database with two parts, the open \textbf{training} set with 3,634 videos, and the reserved \textbf{test} set with 909 videos (\textit{We will maintain a test server for future methods to evaluate on the test set}). Moreover, we evaluate a single-dimension variant for MaxVQA on three common in-the-wild VQA datasets with only overall quality scores available: \textbf{KoNViD-1k} (1200 videos), \textbf{LIVE-VQC} (585 videos), \textbf{YouTube-UGC} (1147 videos). Results reported for existing databases are the mean results of 10 random train-test splits. 

\subsubsection{Baseline Methods for Benchmark Study} To better increase the diverse of our benchmark study, we choose representative state-of-the-art VQA methods with different characteristics:
\begin{itemize}
    \item TLVQM (2019): Representative handcraft \textit{classical} method.
    \item VIDEVAL (2021): Representative handcraft \textit{classical} method.
    \item RAPIQUE (2021): Representative method that combines \textit{deep} CNN features with handcraft \textit{classical} features.
    \item VSFA (2019): The first \textit{deep} CNN-feature-based method.
    \item BVQA-\textit{Li} (2022): An improved \textit{deep} CNN-feature-based method, with an additional \textbf{temporal} 3D-CNN backbone.
    \item FAST-VQA (2022): The first \textit{\textbf{end-to-end} deep} VQA method.
\end{itemize}
and several representative \underline{zero-shot} VQA methods to explore between these objective metrics and the dimensions in the Maxwell:
\begin{itemize}
    \item NIQE (2013): NSS-based \textbf{spatial} zero-shot quality evaluator.
    \item TPQI (2022): State-of-the-art zero-shot \textbf{temporal} VQA method.
    \item SAQI (2023): Zero-shot CLIP-based VQA method with only \textbf{abstract} (\textit{i.e. high/low quality; good/bad}) prompt design.
\end{itemize}

We evaluate all baseline methods with official implementations. 

\begin{table*}
\footnotesize
\caption{Prediction of which dimension is closer to subjective quality scores in \orange{existing databases}? Top-5 dimensions are highlighted with (ranks) in parenthesis.}\label{tab:generalization}
\vspace{-4pt}\setlength\tabcolsep{5.5pt}
\renewcommand\arraystretch{1.25}
\footnotesize
\centering
\vspace{-8pt}
\resizebox{\linewidth}{!}{\begin{tabular}{l|c|c|c|c|c|c|c|c|c|c|c|c|c|c|c|c}
\hline
   Dimension  & \blue{A-1} & \blue{A-2} & \blue{A-3} & \blue{A-4} & \blue{[A-5]} & \blue{A-all} & \bgreen{T-1} & \bgreen{T-2} & \bgreen{T-3} & \bgreen{T-4} & \bgreen{[T-5]} & \bgreen{T-6} & \bgreen{T-7} & \bgreen{[T-8]} & \bgreen{T-all} & \textbf{O}   \\ \hline
  Existing Database          & PLCC    & PLCC   & PLCC   & PLCC   & PLCC   & PLCC   & PLCC   & PLCC   & PLCC   & PLCC   & PLCC   & PLCC   & PLCC  & PLCC & PLCC & PLCC \\ \hline 

LIVE-VQC &0.681&0.708&0.733&0.747&0.777&0.772&0.767&\cellcolor{mred!15}0.821 (3)&0.767&\cellcolor{mred!15}0.821 (3)&0.700&0.765&0.763&\cellcolor{mred!15}0.816 (5)&\cellcolor{mred!15}\textbf{0.830} (1)& \cellcolor{mred!15}0.822 (2) \\ \hdashline
KoNViD-1k & 0.751&0.776&0.759&0.778&0.772&\cellcolor{mred!15}0.825 (4)&\cellcolor{mred!15}0.819 (5)&\cellcolor{mred!15}0.859 (2)&0.813&0.824&0.645&0.768&0.800&0.737&\cellcolor{mred!15}0.859 (2)&\cellcolor{mred!15}\textbf{0.865} (1)\\ \hdashline
YouTube-UGC & 0.681&0.713&0.728&0.756&0.671&0.794&\cellcolor{mred!15}0.819 (4)&\cellcolor{mred!15}0.819 (4)&\cellcolor{mred!15}0.821 (2)&0.793&0.553&0.714&0.814&0.746&\cellcolor{mred!15}0.821 (2)&\cellcolor{mred!15}\textbf{0.830} (1)\\ 
\hline
\end{tabular}}
\vspace{-10pt}
\end{table*}

\begin{table*}
\footnotesize
\caption{Ablation Studies: Effects of Low-level-enhanced Visual Backbone and Contextual Prompts on Maxwell. The metric is PLCC (same as other tables).}\label{tab:ablation}
\vspace{-4pt}\setlength\tabcolsep{5.5pt}
\renewcommand\arraystretch{1.25}
\footnotesize
\centering
\vspace{-8pt}
\resizebox{\linewidth}{!}{\begin{tabular}{l|c|c|c|c|c|c|c|c|c|c|c|c|c|c|c|c}
\hline
   Variants / Dimension  & \blue{A-1} & \blue{A-2} & \blue{A-3} & \blue{A-4} & \blue{[A-5]} & \blue{A-all} & \bgreen{T-1} & \bgreen{T-2} & \bgreen{T-3} & \bgreen{T-4} & \bgreen{[T-5]} & \bgreen{T-6} & \bgreen{T-7} & \bgreen{[T-8]} & \bgreen{T-all} & \textbf{O}   \\ \hline

\textbf{Baseline A}: \textit{Zero-shot} CLIP & 0.143  & 0.322 & 0.411 & -0.028  & 0.192 & 0.153 & 0.498 & 0.414 & 0.058 & 0.241 & 0.346 & 0.254 & 0.174 & 0.053 & 0.218 & 0.467  \\ 
\textbf{A}+$\mathrm{Ctx}$ & 0.551 & 0.598 & 0.637 & 0.662 & 0.582 & 0.569 & 0.746  & 0.664 & 0.627 & 0.662 & 0.591 & 0.670 & 0.634  & 0.414 & 0.634 & 0.670  \\
\textbf{A}+$\mathrm{Ctx}$+MLP (\textit{w/o} $f_\mathcal{V}^\mathrm{FAST}$) & 0.580 & 0.626 & 0.606 & 0.615 & 0.618 & 0.683 & 0.766 & 0.662 & 0.675 & 0.670 & 0.593 & 0.636 & 0.673 & 0.433 & 0.722 & 0.735 \\  \hdashline
\textbf{Baseline B}: FAST-VQA & 0.614 & 0.630 & 0.696 & 0.709 & 0.646 & 0.721 & 0.800 & 0.724 & 0.755 & 0.731 & 0.751 & {0.695} & {0.736} & 0.654 & 0.803  &  0.782 \\ 

\textbf{B}+CLIP visual (\textit{w/o} textual) & 0.644 & 0.678 & 0.699 & 0.723 & 0.698 & 0.765 & 0.815  & 0.734 & 0.772 & 0.748 & 0.767 & 0.733 & 0.743 &  0.648 & 0.809 & 0.802 \\ \hdashline

\textbf{MaxVQA} (all)
&\textbf{0.681}&\textbf{0.701}&\textbf{0.757}&\textbf{0.749}&\textbf{0.712}&\textbf{0.775}&\textbf{0.825}&\textbf{0.748}&\textbf{0.776}&\textbf{0.761}&\textbf{0.782}&\textbf{0.748}&\textbf{0.763}&\textbf{0.684}&\textbf{0.827}&\textbf{0.813} \\
\hline
\end{tabular}}
\vspace{-10pt}
\end{table*}

\subsection{Benchmarking on the Maxwell}
\label{sec:bench}

\subsubsection{Zero-shot Approaches} We first benchmark representative zero-shot quality indices on Maxwell. Specifically, we do not fit the scores with any dimensions, but evaluate how these indices match specific dimensions. The CLIP-based SAQI shows far better performance than NSS-based NIQE, proving the potential of CLIP on VQA. Still, without temporal modeling, it shows notably lower (-25\%) accuracy on two temporal distortions: Flicker (\bgreen{[T-5]}) and Fluency (\bgreen{[T-8]}) than the specific temporal VQA index, TPQI. This suggests that we need to include temporal modeling ability for a CLIP-based approach to better solve in-the-wild VQA problem.
\subsubsection{Existing Supervised Methods, and MaxVQA} We also benchmark existing supervised methods with a multi-task training strategy. Classical methods even struggle to surpass zero-shot approaches on certain dimensions. For deep methods, while FAST-VQA reaches champions on overall performance and most dimensions, BVQA-\textit{Li} is more competitive on temporal-related dimensions (\textbf{A-5, T-5, T-8}). The Maxwell benchmark can then suggest that an independent temporal backbone may improve temporal modeling in VQA, besides only concluding that FAST-VQA is more effective. For the proposed \textbf{MaxVQA}, with CLIP-visual features and language-prompted modeling, it can further notably outperform FAST-VQA especially on semantic-related dimensions (\textbf{A-x}, +9.4\% in-average). The results suggest that accurately assessing multiple quality factors is a more challenging task than only predicting overall quality scores.

\subsection{Evaluation on Existing Databases} 
\label{sec:existing}

 \subsubsection{Training and Testing on Existing Databases} We also train and evaluate the proposed language-prompted MaxVQA on existing VQA databases, with a variant with only single objective on overall score (\textbf{O}). As compared in Table~\ref{tab:existing}, the proposed MaxVQA can significantly outperform baseline methods on all datasets, proving the robust excellent performance of its visual-language-based design.

 \subsubsection{Analyzing Existing Databases with Multi-axis Predictions} In Table~\ref{tab:generalization}, we compare between multi-dimensional MaxVQA predictions trained with Maxwell and overall scores in existing datasets. First, we prove that MaxVQA can generalize well from Maxwell to existing datasets on overall quality (\textbf{O}) prediction. Furthermore, we observe that unlike KoNViD-1k and YouTube-UGC, the subjective scores in LIVE-VQC are more correlated with predictions of the technical perspective (\bgreen{T-all}) than overall quality (\textbf{O}). This result suggests that subjective studies conducted for different databases might not follow the same scoring standards implicitly. Among the specific factors, we notice that Focus (T-2) is important for all databases, while LIVE-VQC is especially concerned about Motion Blur (T-4) and Fluency (T-8) as all videos are from hand-held smartphones. Moreover, KoNViD-1k is obviously more concerned about pure aesthetic predictions (A-all) and contents (A-1) than others.

\subsection{Ablation Studies}
\label{sec:abl}

 We conduct ablation studies to investigate the effects of proposed modules in MaxVQA, as listed in Table~\ref{tab:ablation}. Zero-shot CLIP generally performs poorly on specific quality factors, validating our second claim in Sec.~\ref{sec:modifyclip} and the necessity of learnable text prompts.

\subsubsection{Effects of Enhanced Visual Backbone} Considering the context-prompted CLIP as baseline, we discuss the effects of the proposed enhanced visual backbone. Without the FAST-VQA features, either the original or adapted~\cite{clipadapter} CLIP features lead to less accuracy, especially for \textit{temporal distortions} ([T-5] Flicker and [T-8] Fluency), where the FAST-VQA features can improve \textbf{58\%} and \textbf{32\%}, proving the vitality of including temporal modeling upon baseline CLIP.

\subsubsection{Effects of CLIP} From another perspective, we treat the FAST-VQA as our baseline method, and investigate the effects of pre-trained CLIP in the MaxVQA. Directly integrating CLIP visual features can lead to significant improvements, especially on aesthetic dimensions. The text-prompted modeling further notably improve the performance than the naive multi-task variant \textit{without} textual modeling, proving the rationality of our multi-modal design.


\begin{figure}
    \centering
    \includegraphics[width=\linewidth]{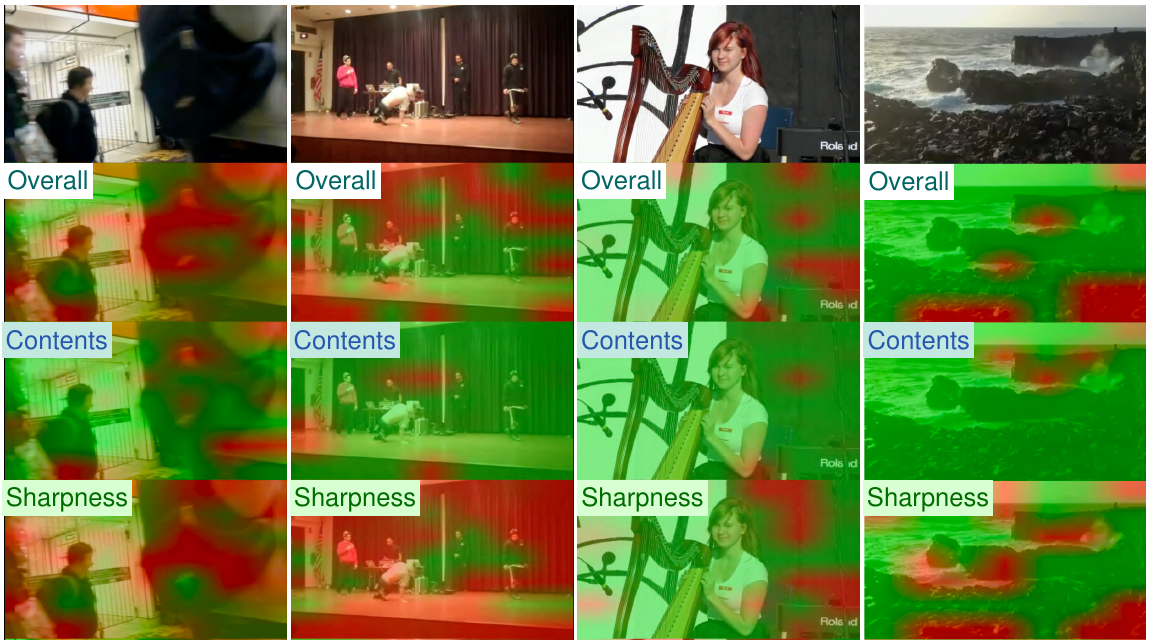}
    \vspace{-18pt}
    \caption{Multi-axis local quality maps for dimensions Overall (O), Contents (A-1) and Sharpness (T-1) of the proposed MaxVQA, showing their differences.}
    \label{fig:localmaps}
    \vspace{-15pt}
\end{figure}

\subsection{Multi-Axis Local Quality Maps}
\label{sec:qual}

As $f_{V_t}^\text{Final}$ are localized features, we are able to generate local quality maps (similar as \cite{fastvqa}) from different dimensions, so as to qualitatively examine the their quality concerns. As illustrated in Fig.~\ref{fig:localmaps}, the proposed MaxVQA can comprehensively detect \textit{degraded} or \textit{unappealing} local areas as worse overall quality (O); moreover, the Contents (A-1) axis predicts that human-related areas in videos are with better quality than backgrounds; and the Sharpness (T-1) axis can well-distinguish fuzzy areas (\textit{e.g. backpack in the leftmost video}), suggesting that MaxVQA can distinguish differences among axes.

\section{Conclusion}

Our study significantly expands the scope of in-the-wild Video Quality Assessment (VQA) by explaining subjective quality scores with specific factors. A large-scale in-the-wild VQA database, named \textbf{Maxwell}, is created to gather more than two million human opinions across 13 specific quality-related factors, including technical distortions \textit{e.g. noise, flicker} and aesthetic factors \textit{e.g. contents}. With the Maxwell database, we investigate the relationships between various perceptual factors and examine how they influence overall quality opinions. Moreover, the Maxwell establishes a novel multi-dimensional benchmark for objective VQA methods to assess their strengths and weaknesses in capturing various quality issues, providing more detailed guidance for future methods. Additionally, the study introduces the \textbf{MaxVQA}, a language-prompted VQA approach that can jointly evaluate multiple specific quality factors and overall perceptual quality scores, achieving state-of-the-art results with excellent generalization abilities. We hope that our efforts will bring along new insights and advancements in the VQA field.

\section{Acknowledgements}

This study is supported under the RIE2020 Industry Alignment Fund – Industry Collaboration Projects (IAF-ICP) Funding Initiative, as well as cash and in-kind contribution from the industry partner(s).

\appendix

\section{Extended Analyses for Maxwell}

\subsection{Distributions}

\begin{figure}
    \centering
    \includegraphics[width=\linewidth]{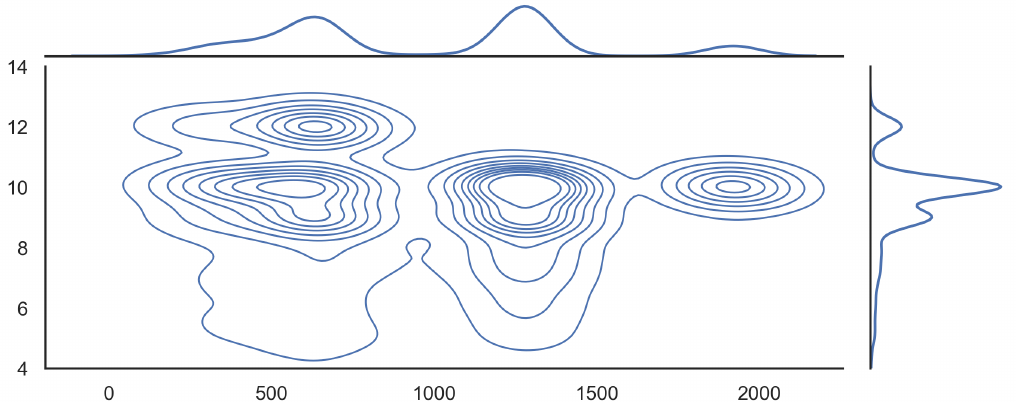}
    \vspace{-16pt}
    \caption{Joint distributions of video resolution and duration in Maxwell.}
    \label{fig:disrd}
\end{figure}

\subsubsection{Distributions of Video Resolution and Duration}

In Fig.~\ref{fig:disrd}, we show the joint distributions of video resolution (\textit{long edge, or max value between height and width}) and video duration (\textit{number of frames divided by framerate}), showing that the videos collected in the Maxwell database are in multiple resolution, and with various length, demonstrating the diversity of the constructed database.

\begin{figure}
    \centering
    \includegraphics[width=\linewidth]{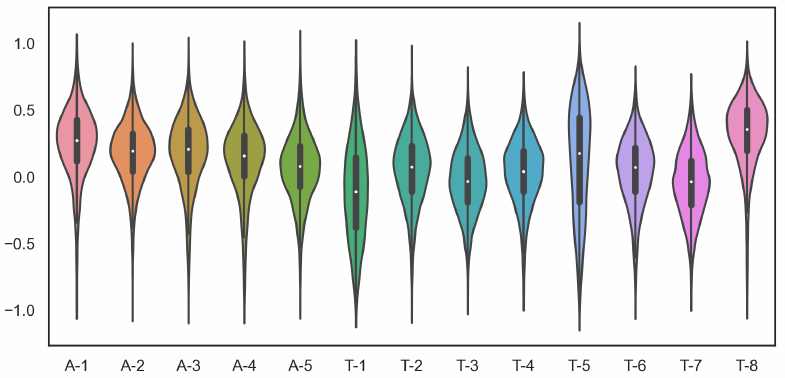}
    \vspace{-16pt}
    \caption{Distributions of mean factor opinion ($\mathrm{MOS}_a$) in each dimension $a$.}
    \vspace{-15pt}
    \label{fig:disss}
\end{figure}

\subsubsection{Distributions of Factor Scores}
\label{sec:distfs} In addition to the absolute responses as studied in the main paper, we also measure the distributions of scores in each dimension, as illustrated in Fig.~\ref{fig:disss}. For the four factors which are with especially high absolute responses ((A-1) Contents, (T-1) Sharpness, [T-5] Flicker, [T-8] Fluency), we notice that contents and fluency have obvious positive-biased opinions, while the scores for sharpness and flicker dimensions show very strong skewness (which should be easier to be distinguished).

\section{More Details for MaxVQA}

\begin{table*}
\footnotesize
\caption{Generalization evaluations among existing in-the-wild VQA databases. }\label{tab:genexisting}
\vspace{-4pt}\setlength\tabcolsep{7pt}
\renewcommand\arraystretch{1.25}
\footnotesize
\centering
\vspace{-8pt}
\resizebox{\linewidth}{!}{\begin{tabular}{l|cc|cc|cc|cc|cc|cc}
\hline
\textbf{Training Set} & \multicolumn{4}{c|}{{KoNViD-1k}} & \multicolumn{4}{c|}{{LIVE-VQC}} & \multicolumn{4}{c}{{Youtube-UGC}} \\ \hline
\textbf{Test Set} & \multicolumn{2}{c|}{{LIVE-VQC}} & \multicolumn{2}{c|}{{Youtube-UGC}}  & \multicolumn{2}{c|}{{KoNViD-1k}} & \multicolumn{2}{c|}{{Youtube-UGC}}   & \multicolumn{2}{c|}{{LIVE-VQC}} & \multicolumn{2}{c}{{KoNViD-1k}} \\ \hline
\textit{}                         & SRCC$\uparrow$                        & PLCC$\uparrow$                        & SRCC$\uparrow$                        & PLCC$\uparrow$                 & SRCC$\uparrow$                        & PLCC$\uparrow$                        & SRCC$\uparrow$                        & PLCC$\uparrow$        & SRCC$\uparrow$                        & PLCC$\uparrow$               & SRCC$\uparrow$                       & PLCC$\uparrow$                                         \\ 
\hline
TLVQM (2019, TIP)\cite{tlvqm} & 0.573 & 0.629 & 0.354 & 0.378 & 0.640  & 0.630& 0.218 & 0.250 & 0.488 & 0.546 & 0.556 & 0.578 \\
CNN-TLVQM (2020, MM)\cite{cnntlvqm}             & 0.713                      & 0.752    & {0.424} &   {0.469}               & 0.642 & 0.631 & 0.329 & 0.367      & 0.551 & 0.578 & 0.588 & 0.619                               \\ 

VIDEVAL (2021, TIP)\cite{videval}                    & 0.627                         & 0.654   & 0.370 & 0.390                  & 0.625                        & 0.621 & 0.302 & 0.318  & 0.542                         & 0.553      & 0.610 &   0.620                           \\ 
MDTVSFA (2021, IJCV)\cite{mdtvsfa}                     & {0.716}                        & {0.759}    & {0.408} &  {0.443}                 & 0.706                         & {0.711}  & {0.355} &  {0.388}      & {0.582}                        & {0.603}   & {0.649} &   {0.646}                     \\ 
GST-VQA (2022, TCSVT)\cite{gstvqa}             & 0.700                      & 0.733    & \gray{NA} &   \gray{NA}               & {0.709} & 0.707 & \gray{NA} & \gray{NA}       & \gray{NA}                        & \gray{NA}             & \gray{NA}  &   \gray{NA}                                 \\ 
BVQA-\textit{Li} (2022, TCSVT)\cite{bvqa2021} &   {0.695} & {0.712}       & {0.780}     & {0.780}         & {0.738} & {0.721}       & {0.602}     & {0.602} &  {0.689}     & {0.727} & {0.785} & {0.782}                       \\
DisCoVQA (2023, TCSVT)\cite{discovqa}           & {0.782}               & {0.797}       & {0.415} & {0.449}    & {0.792}               & {0.785}  & {0.409} & 0.432   & 0.661               & 0.685      &  0.686 &   0.697               \\ \hdashline

\textbf{MaxVQA} (Ours) & \textbf{0.793} & \textbf{0.832} & \textbf{0.867} & \textbf{0.857} & \textbf{0.833} & \textbf{0.831} & \textbf{0.846} & \textbf{0.824} & \textbf{0.804} & \textbf{0.812} & \textbf{0.855} & \textbf{0.852}\\
\hline
\end{tabular}}
\vspace{-10pt}
\end{table*}

\begin{table*}
\footnotesize
\caption{Ablation Studies: Effects of Low-level-enhanced Visual Backbone and Contextual Prompts on Maxwell. }\label{tab:ablationp}
\vspace{-4pt}\setlength\tabcolsep{5.5pt}
\renewcommand\arraystretch{1.25}
\footnotesize
\centering
\vspace{-8pt}
\resizebox{\linewidth}{!}{\begin{tabular}{l|c|c|c|c|c|c|c|c|c|c|c|c|c|c|c|c}
\hline
   Objective  & \blue{A-1} & \blue{A-2} & \blue{A-3} & \blue{A-4} & \blue{[A-5]} & \blue{A-all} & \bgreen{T-1} & \bgreen{T-2} & \bgreen{T-3} & \bgreen{T-4} & \bgreen{[T-5]} & \bgreen{T-6} & \bgreen{T-7} & \bgreen{[T-8]} & \bgreen{T-all} & \textbf{O}   \\ \hline
  Variants          & PLCC    & PLCC   & PLCC   & PLCC   & PLCC   & PLCC   & PLCC   & PLCC   & PLCC   & PLCC   & PLCC   & PLCC   & PLCC  & PLCC & PLCC & PLCC \\ \hline

Shared Initial Prompt & 0.578 & 0.633 & 0.619 & 0.670 & 0.656 & 0.741 & 0.765 & 0.730 & 0.742 & 0.739 & 0.655 & 0.688 & 0.728 & 0.571 & 0.820 & 0.811 \\

\textbf{MaxVQA} (Axis-specific Prompt)
&\textbf{0.681}&\textbf{0.701}&\textbf{0.757}&\textbf{0.749}&\textbf{0.712}&\textbf{0.775}&\textbf{0.825}&\textbf{0.748}&\textbf{0.776}&\textbf{0.761}&\textbf{0.782}&\textbf{0.748}&\textbf{0.763}&\textbf{0.684}&\textbf{0.827}&\textbf{0.813} \\
\hline
\end{tabular}}
\end{table*}

\subsection{Fragment Sampling}

We follow the original fragment sampling strategy as proposed by FAST-VQA~\cite{fastvqa}, introduced as follows. For the first step, we cut the $t$-th video frame $\mathcal{V}_t$ into $G_f\times G_f$ uniform grids with the same sizes, denoted as $\mathcal{G}_t=\{g_t^{0,0},..g_t^{i,j},..g_t^{G_f-1,G_f-1}\}$, where $g_t^{i,j}$ denotes the grid 
in the $i$-th row and $j$-th column. The uniform grid partition process is formalized as follows.
\begin{equation}
    g_t^{i,j} = \mathcal{V}_t[\frac{i\times H}{G_f}:\frac{(i+1)\times H}{G_f},\frac{j\times W}{G_f}:\frac{(j+1)\times W}{G_f}]\label{eq:1}
\end{equation}
where $H$ and $W$ denote the height and width of the video frame. Then, we employ random patch sampling to select one mini-patch $\mathcal{MP}_t^{i,j}$ of size of $S_f \times S_f$ from each grid $g_t^{i,j}$, as follows:
\begin{equation}
     \mathcal{MP}_{t}^{i,j} = \mathbf{S}_{t}^{i,j}(g_{t}^{i,j})\label{eq:2}
\end{equation}
where $\mathbf{S}_{t}^{i,j}$ is the patch sampling operation for frame $t$ and grid $i,j$.

The sampling operation ($\mathbf{S}$) is temporally aligned to identify the temporal distortions (\textit{e.g.} \bgreen{[T-5]} Flicker and \bgreen{[T-8]} Fluency).

\begin{equation}
     \mathbf{S}_{t}^{i,j} = \mathbf{S}_{\hat{t}}^{i,j}~~~~~~\forall~0\leq t,\hat{t}<T,~ 0\leq i, j < G_f
\end{equation}

The mini-patches are then spliced into their original positions:
\begin{equation}
\begin{aligned}
        \mathcal{F}_t^{i,j}&= \mathcal{F}_t[i\times S_f:(i+1)\times S_f, j\times S_f:(j+1)\times S_f] \\&= \mathcal{MP}_t^{i,j},~~~~~~~~~~~~~~~~~~~~ 0\leq i, j < G_f \label{eq:3} 
\end{aligned}
\end{equation}
where $\mathcal{F}$ denote the spliced and temporally aligned mini-patches after the fragment sampling, named as fragments.

\subsection{Details for Feature Fusion Module}

In the Feature Fusion Module, we concatenate the CLIP-visual features with FAST-VQA features and feed the concatenated features into a patch-wise MLP for feature fusion. Specificall the Video Swin Transformer backbone, the FAST-VQA feature pixel for mini-patch $\mathcal{MP}_{t}^{i,j}$ can be denoted as $(F_{\mathcal{V}}^\text{FAST})_{t,i,j}$. For the CLIP-visual features, as we have extracted local features after the attention pooling, the feature pixel at the same location can also be obtained from $(F_{V,t}^\text{Local})_{i,j}$. Then, denote the fully-connected layers as $\mathrm{FC_1}, \mathrm{FC_2}$, GELU activation function as $\mathrm{Act}$, dropout layer as $\mathrm{Drop}$, the feature fusion module calculates the residual-wise modification to the original $F_{V,t}^\text{Local}$ in the CLIP cross-modality representation space:
\begin{align}
    (F_{V,t}^\text{Res})_{i,j} &= \mathrm{FC_2}(\mathrm{Drop}(\mathrm{Act}(\mathrm{FC_1}([(F_{\mathcal{V}}^\text{FAST})_{t,i,j}, (F_{V,t}^\text{Local})_{i,j}])))) \\
    (F_{V,t}^\text{Final})_{i,j} &= (F_{V,t}^\text{Res})_{i,j} + (F_{V,t}^\text{Local})_{i,j}
\end{align}
where $[a,b]$ denotes connecting tensors $a,b$ in channel dimension.

\section{Extended Results of MaxVQA}
\subsection{Quantitative Results}

\begin{table}
\footnotesize
\caption{Mix-dataset Training on LIVE-VQC, KoNViD-1k and YouTube-UGC.}\label{tab:mix}
\vspace{-4pt}\setlength\tabcolsep{4.5pt}
\renewcommand\arraystretch{1.25}
\footnotesize
\centering
\vspace{-8pt}
\resizebox{\linewidth}{!}{\begin{tabular}{l|cc|cc|cc}
\hline
   Dataset  & \multicolumn{2}{c|}{{LIVE-VQC}}   & \multicolumn{2}{c|}{{KoNViD-1k}}    & \multicolumn{2}{c}{{YouTube-UGC}}      \\ \hline
  Methods           & SRCC$\uparrow$& PLCC$\uparrow$  & SRCC$\uparrow$& PLCC$\uparrow$         & SRCC$\uparrow$& PLCC$\uparrow$       \\ \hline 
{FAST-VQA}\cite{fastvqa} (Separate Training) & 0.849 & 0.862 & 0.891 & 0.892 & 0.855 & 0.852 \\
\textbf{MaxVQA} (Separate Training) & {0.854} & {0.873}  & {0.894} &  {0.895}  & {0.894} & {0.890} \\ \hdashline
{FAST-VQA}\cite{fastvqa} (Mixed Training) & 0.831 & 0.850 & 0.873 & 0.873 & 0.841 & 0.837 \\
\textbf{MaxVQA} (Mixed Training) & {0.851} & {0.872}  & {0.883} &  {0.887}  & {0.887} & {0.882} \\
\hline
\end{tabular}}
\end{table}

\begin{figure}
    \centering
    \includegraphics[width=0.98\linewidth]{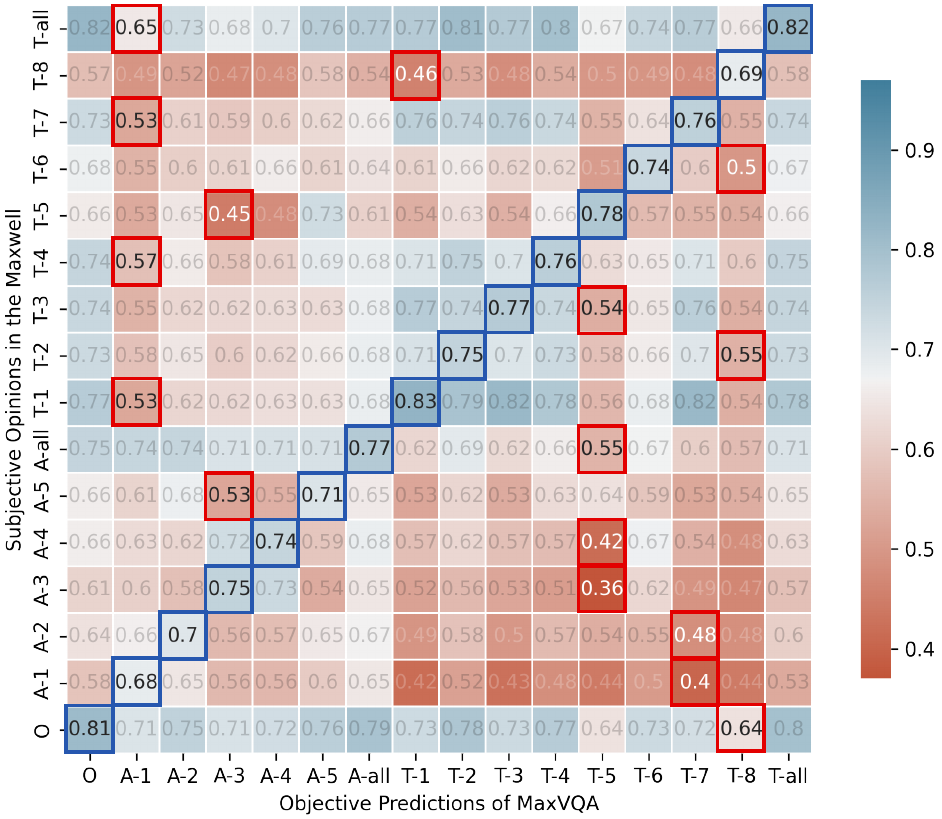}
    \vspace{-10pt}
    \caption{The cross-dimension validation PLCC between MaxVQA objective predictions on different dimensions and subjective opinions in the Maxwell database. \rblue{Blue} boxes for the most correlated prediction with respective to each subjective dimension; \bred{red} boxes for the least correlated ones.}
    \label{fig:corrvqa}
    \vspace{-15pt}
\end{figure}

\begin{figure*}
    \centering
    \includegraphics[width=\linewidth]{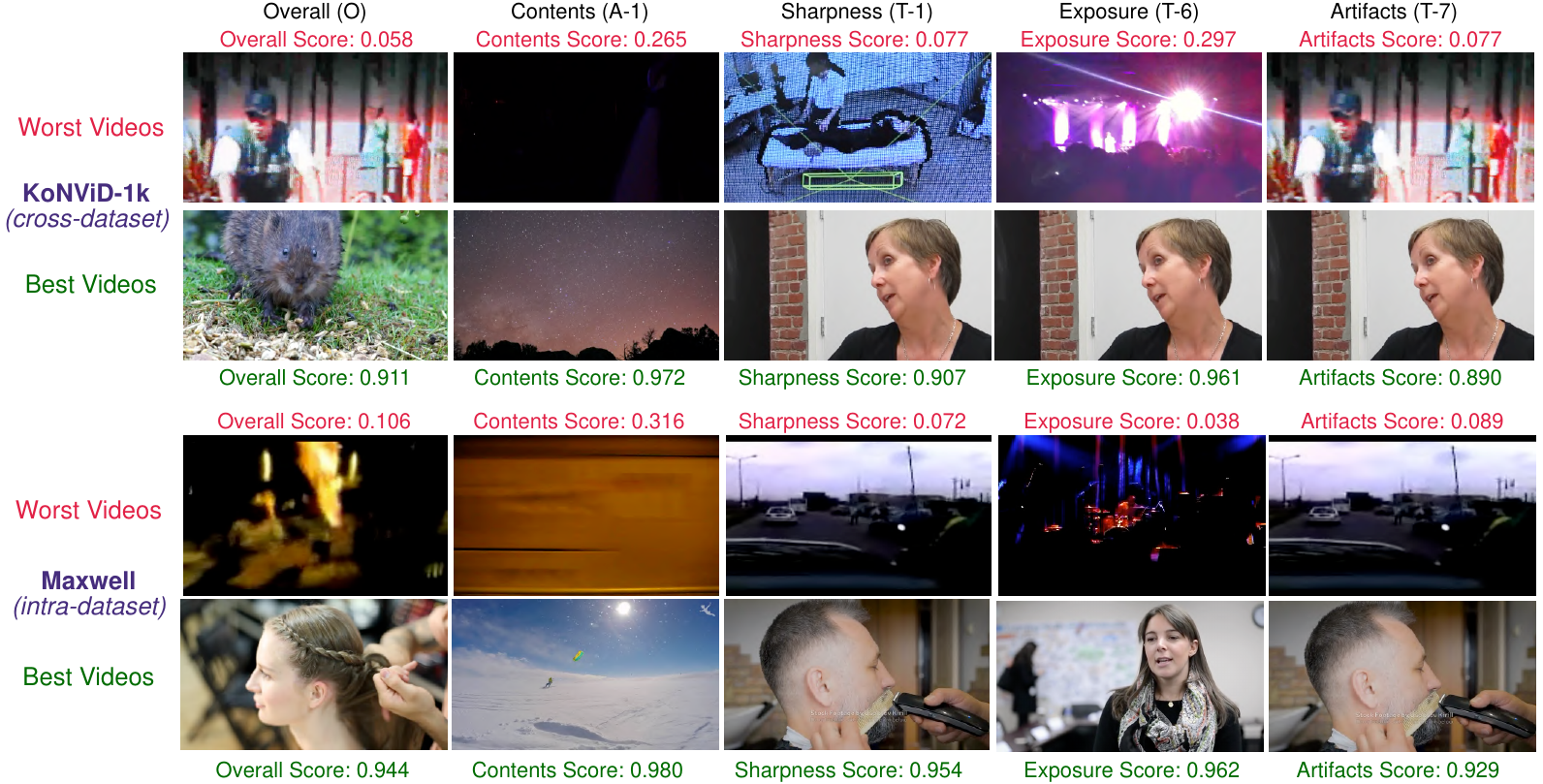}
    \vspace{-20pt}
    \caption{Videos with best and worst \underline{MaxVQA-predicted} scores in different \textit{spatial} dimensions in KoNViD-1k and Maxwell test set.}
    \label{fig:bws}
\end{figure*}
\begin{figure*}
\vspace{-10pt}
    \centering
    \includegraphics[width=\linewidth]{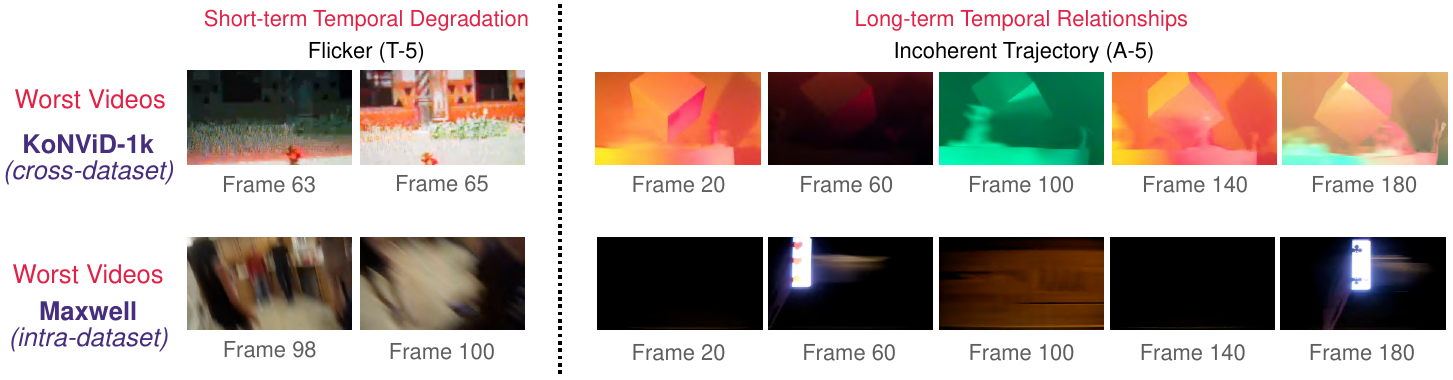}
    \vspace{-20pt}
    \caption{Videos with worst \underline{MaxVQA-predicted} scores in different \textit{temporal} dimensions in KoNViD-1k and Maxwell test set.}
    \label{fig:wt}
    \vspace{-8pt}
\end{figure*}

\subsubsection{Unifying Datasets: Cross-dataset and Mix-dataset Results}

In addition to evaluating the generalization ability of MaxVQA trained on MaxWell database, we also evaluate its cross-dataset performance among existing VQA databases, as shown in Table~\ref{tab:genexisting}. As illustrated in the table, with strong quality-sensitive features and robust language-prompted modeling, the proposed MaxVQA can also reach unprecedented generalization ability among existing VQA databases. Moreover, with the training procedures as defined in MDTVSFA~\cite{mdtvsfa}, we evaluate the mix-dataset training ability of the proposed MaxVQA, which is similar with separate training (\textit{-0.7\% when mixed}) and is more stable than FAST-VQA (\textit{-1.8\% when mixed}). All these results suggest that the proposed MaxVQA moves a step forward towards unified in-the-wild VQA that one pre-trained model can robustly predict for various scenarios.

\subsubsection{Design of Initial Text Prompts} 

In MaxVQA, we design the initial text prompts differently for each dimension, so as to achieve multi-objective training without any dimension-specific parameters, and prove its superiority than naive multi-task with multiple regression heads. Moreover, in Table~\ref{tab:ablationp}, we further prove the necessity to set different prompts for different dimensions, where a shared initial prompt (\textit{all set as ``high quality/low quality''}) cannot predict well on the specific factors, especially the factors with lower correlation to the overall quality scores (\textit{e.g.} (A-1) Contents).

\subsection{Qualitative Results}

\subsubsection{Cross-dimension Validation} In Fig. 2 of main paper, we analyzed the correlations among subjective opinions for different dimensions. In Fig.~\ref{fig:corrvqa}, we further measure the cross-dimension validation results, \textit{i.e.} the correlations between the objective predictions of MaxVQA in a certain dimension with subjective opinions in all dimensions. From the cross-validation results, we first confirm that for each single dimension, the prediction on the respective dimension best correlates with the subjective opinions in it, proving the general effectiveness of the proposed MaxVQA. Moreover, we notice that the technical quality predictions \textbf{(T-x)} are especially less correlated with subjective aesthetic opinions, and vice versa, further proving the rationality of separately considering the two perspectives. Still, we notice that some technical factors ((T-1) Sharpness, (T-3) Noise, (T-7) Artifacts) are not so effectively distinguished with the current model, perhaps because they are likely to exist together in real-world in-the-wild videos. We plan to introduce semi-supervised training together with synthetic distortions to further improve the disentanglement among these factors.

\subsubsection{Best and Worst Videos in Different Dimensions} Furthermore, we show the best and worst videos with perspective of different dimensions in the following databases: test set of Maxwell (\textit{intra-dataset}) and KoNViD-1k (\textit{cross-dataset}). Similar as main paper Fig.~7, we choose overall quality (O), contents (A-1), sharpness (T-1), artifacts (T-7) and exposure (T-6) for visualization, as illustrated in Fig.~\ref{fig:bws}. While the overall score (O) provides comprehensive evaluation on the video's quality, the predictions on the technical factors can effectively identify the videos with extremely low sharpness, over or under exposure, or compression artifacts. Moreover, as suggested in Sec.~\ref{sec:distfs}, the content dimension usually results in positive feelings of subjects. This can also be well captured by MaxVQA, which can effectively pick videos with very delicate contents in datasets. We also notice that the high cross-relation between sharpness (T-1) and artifacts (T-7) are mostly due to their high statistical relevance (\textit{fuzzy videos are usually with strong artifacts; vice versa}).

In Fig.~\ref{fig:wt}, we further discuss the two-level temporal quality, the aesthetic-related consistency on trajectory (A-5) which happens in a longer term, and the technical distortion flicker (T-5) that usually happens in a very short term. As illustrated in the figure, the MaxVQA can identify the inter-frame abrupt changes in the Flicker dimension, and notice the inconsistent long-term relationships (\textit{appearance level and semantic level}) during the whole video with Trajectory dimension. The results suggest that MaxVQA can comprehensively evaluate temporal quality, which is demonstrated an important part of in-the-wild VQA in our subjective studies.

{
\bibliographystyle{acm}
\bibliography{egbib}

\begin{thebibliography}{10}

\bibitem{itu}
Recommendation 500-10: Methodology for the subjective assessment of the quality
  of television pictures.
\newblock ITU-R Rec. BT.500, 2000.

\bibitem{msu}
{\sc Antsiferova, A., Lavrushkin, S., Smirnov, M., Gushchin, A., Vatolin,
  D.~S., and Kulikov, D.}
\newblock Video compression dataset and benchmark of learning-based
  video-quality metrics.
\newblock In {\em Thirty-sixth Conference on Neural Information Processing
  Systems Datasets and Benchmarks Track\/} (2022).

\bibitem{basicvsr}
{\sc Chan, K.~C., Wang, X., Yu, K., Dong, C., and Loy, C.~C.}
\newblock Basicvsr: The search for essential components in video
  super-resolution and beyond.
\newblock In {\em CVPR\/} (2021).

\bibitem{gstvqa}
{\sc Chen, B., Zhu, L., Li, G., Lu, F., Fan, H., and Wang, S.}
\newblock Learning generalized spatial-temporal deep feature representation for
  no-reference video quality assessment.
\newblock {\em IEEE TCSVT\/} (2021).

\bibitem{rirnet}
{\sc Chen, P., Li, L., Ma, L., Wu, J., and Shi, G.}
\newblock Rirnet: Recurrent-in-recurrent network for video quality assessment.
\newblock {\em ACM MM\/} (2020).

\bibitem{lightvqa}
{\sc Dong, Y., Liu, X., Gao, Y., Zhou, X., Tan, T., and Zhai, G.}
\newblock Light-vqa: A multi-dimensional quality assessment model for low-light
  video enhancement.
\newblock In {\em n Proceedings of the 31st ACM International Conference on
  Multimedia\/} (2023).

\bibitem{vit}
{\sc Dosovitskiy, A., Beyer, L., Kolesnikov, A., Weissenborn, D., Zhai, X.,
  Unterthiner, T., Dehghani, M., Minderer, M., Heigold, G., Gelly, S., et~al.}
\newblock An image is worth 16x16 words: Transformers for image recognition at
  scale.
\newblock {\em arXiv preprint arXiv:2010.11929\/} (2020).

\bibitem{spaq}
{\sc Fang, Y., Zhu, H., Zeng, Y., Ma, K., and Wang, Z.}
\newblock Perceptual quality assessment of smartphone photography.
\newblock In {\em CVPR}.

\bibitem{clipadapter}
{\sc Gao, P., Geng, S., Zhang, R., Ma, T., Fang, R., Zhang, Y., Li, H., and
  Qiao, Y.}
\newblock Clip-adapter: Better vision-language models with feature adapters.
\newblock {\em arXiv preprint arXiv:2110.04544\/} (2021).

\bibitem{qualcomm}
{\sc Ghadiyaram, D., Pan, J., Bovik, A.~C., Moorthy, A.~K., Panda, P., and
  Yang, K.-C.}
\newblock In-capture mobile video distortions: A study of subjective behavior
  and objective algorithms.
\newblock {\em IEEE TCSVT 28}, 9 (2018), 2061--2077.

\bibitem{he2016residual}
{\sc He, K., Zhang, X., Ren, S., and Sun, J.}
\newblock Deep residual learning for image recognition.
\newblock In {\em CVPR\/} (2016), pp.~770--778.

\bibitem{kv1k}
{\sc Hosu, V., Hahn, F., Jenadeleh, M., Lin, H., Men, H., Szirányi, T., Li,
  S., and Saupe, D.}
\newblock The konstanz natural video database (konvid-1k).
\newblock In {\em QoMEX\/} (2017), pp.~1--6.

\bibitem{koniq}
{\sc Hosu, V., Lin, H., Sziranyi, T., and Saupe, D.}
\newblock Koniq-10k: An ecologically valid database for deep learning of blind
  image quality assessment.
\newblock {\em IEEE TIP 29\/} (2020), 4041--4056.

\bibitem{distilliaa}
{\sc Hou, J., Ding, H., Lin, W., Liu, W., and Fang, Y.}
\newblock Distilling knowledge from object classification to aesthetics
  assessment.
\newblock {\em IEEE TCSVT\/} (2022).

\bibitem{clipiaa}
{\sc Hou, J., Lin, W., Fang, Y., Wu, H., Chen, C., Liao, L., and Liu, W.}
\newblock Towards transparent deep image aesthetics assessment with tag-based
  content descriptors.
\newblock {\em IEEE Transactions on Image Processing\/} (2023).

\bibitem{objiaa}
{\sc Hou, J., Yang, S., and Lin, W.}
\newblock Object-level attention for aesthetic rating distribution prediction.
\newblock In {\em ACM MM\/} (2020), p.~816–824.

\bibitem{rife}
{\sc Huang, Z., Zhang, T., Heng, W., Shi, B., and Zhou, S.}
\newblock Real-time intermediate flow estimation for video frame interpolation.
\newblock In {\em Proceedings of the European Conference on Computer Vision
  (ECCV)\/} (2022).

\bibitem{openclip}
{\sc Ilharco, G., Wortsman, M., Wightman, R., Gordon, C., Carlini, N., Taori,
  R., Dave, A., Shankar, V., Namkoong, H., Miller, J., Hajishirzi, H., Farhadi,
  A., and Schmidt, L.}
\newblock {OpenCLIP}, July 2021.

\bibitem{align}
{\sc Jia, C., Yang, Y., Xia, Y., Chen, Y.-T., Parekh, Z., Pham, H., Le, Q.~V.,
  Sung, Y., Li, Z., and Duerig, T.}
\newblock Scaling up visual and vision-language representation learning with
  noisy text supervision.
\newblock In {\em ICML\/} (2021).

\bibitem{k400data}
{\sc Kay, W., Carreira, J., Simonyan, K., Zhang, B., Hillier, C.,
  Vijayanarasimhan, S., Viola, F., Green, T., Back, T., Natsev, A., Suleyman,
  M., and Zisserman, A.}
\newblock The kinetics human action video dataset.
\newblock {\em ArXiv abs/1705.06950\/} (2017).

\bibitem{vila}
{\sc Ke, J., Ye, K., Yu, J., Wu, Y., Milanfar, P., and Yang, F.}
\newblock Vila: Learning image aesthetics from user comments with
  vision-language pretraining, 2023.

\bibitem{tlvqm}
{\sc Korhonen, J.}
\newblock Two-level approach for no-reference consumer video quality
  assessment.
\newblock {\em IEEE TIP 28}, 12 (2019), 5923--5938.

\bibitem{cnntlvqm}
{\sc Korhonen, J., Su, Y., and You, J.}
\newblock Blind natural video quality prediction via statistical temporal
  features and deep spatial features.
\newblock In {\em ACM MM\/} (2020), p.~3311–3319.

\bibitem{stablevqa}
{\sc Kou, T., Liu, X., Jia, J., Sun, W., Zhai, G., and Liu, N.}
\newblock Stablevqa: A deep no-reference quality assessment model for video
  stability.
\newblock In {\em Proceedings of the 31st ACM International Conference on
  Multimedia\/} (2023).

\bibitem{3dstabilize}
{\sc Lee, Y.-C., Tseng, K.-W., Chen, Y.-T., Chen, C.-C., Chen, C.-S., and Hung,
  Y.-P.}
\newblock 3d video stabilization with depth estimation by cnn-based
  optimization.
\newblock In {\em Proceedings of the IEEE/CVF Conference on Computer Vision and
  Pattern Recognition (CVPR)\/} (June 2021), pp.~10621--10630.

\bibitem{bvqa2021}
{\sc Li, B., Zhang, W., Tian, M., Zhai, G., and Wang, X.}
\newblock Blindly assess quality of in-the-wild videos via quality-aware
  pre-training and motion perception.
\newblock {\em IEEE TCSVT\/} (2022).

\bibitem{agiqa}
{\sc Li, C., Zhang, Z., Wu, H., Sun, W., Min, X., Liu, X., Zhai, G., and Lin,
  W.}
\newblock Agiqa-3k: An open database for ai-generated image quality assessment,
  2023.

\bibitem{vsfa}
{\sc Li, D., Jiang, T., and Jiang, M.}
\newblock Quality assessment of in-the-wild videos.
\newblock In {\em ACM MM\/} (2019), p.~2351–2359.

\bibitem{mdtvsfa}
{\sc Li, D., Jiang, T., and Jiang, M.}
\newblock Unified quality assessment of in-the-wild videos with mixed datasets
  training.
\newblock {\em International Journal of Computer Vision 129}, 4 (2021).

\bibitem{tpqi}
{\sc Liao, L., Xu, K., Wu, H., Chen, C., Sun, W., Yan, Q., and Lin, W.}
\newblock Exploring the effectiveness of video perceptual representation in
  blind video quality assessment.
\newblock In {\em ACM MM\/} (2022).

\bibitem{rankiqa}
{\sc {Liu}, X., {Van De Weijer}, J., and {Bagdanov}, A.~D.}
\newblock Exploiting unlabeled data in cnns by self-supervised learning to
  rank.
\newblock {\em IEEE TPAMI\/} (2019), 1--1.

\bibitem{swin3d}
{\sc Liu, Z., Ning, J., Cao, Y., Wei, Y., Zhang, Z., Lin, S., and Hu, H.}
\newblock Video swin transformer.
\newblock In {\em CVPR\/} (2022).

\bibitem{niqe}
{\sc Mittal, A., Soundararajan, R., and Bovik, A.~C.}
\newblock Making a “completely blind” image quality analyzer.
\newblock {\em IEEE Signal Processing Letters 20}, 3 (2013), 209--212.

\bibitem{avaiaa}
{\sc Murray, N., Marchesotti, L., and Perronnin, F.}
\newblock Ava: A large-scale database for aesthetic visual analysis.
\newblock In {\em CVPR\/} (2012), pp.~2408--2415.

\bibitem{xclip}
{\sc Ni, B., Peng, H., Chen, M., Zhang, S., Meng, G., Fu, J., Xiang, S., and
  Ling, H.}
\newblock Expanding language-image pretrained models for general video
  recognition.
\newblock {\em ECCV\/} (2022).

\bibitem{cvd}
{\sc Nuutinen, M., Virtanen, T., Vaahteranoksa, M., Vuori, T., Oittinen, P.,
  and Häkkinen, J.}
\newblock Cvd2014—a database for evaluating no-reference video quality
  assessment algorithms.
\newblock {\em IEEE TIP 25}, 7 (2016).

\bibitem{gopro}
{\sc Park, D., Kim, J., and Chun, S.~Y.}
\newblock Down-scaling with learned kernels in multi-scale deep neural networks
  for non-uniform single image deblurring.
\newblock {\em arXiv preprint arXiv:1903.10157\/} (2019).

\bibitem{clip}
{\sc Radford, A., Kim, J.~W., Hallacy, C., Ramesh, A., Goh, G., Agarwal, S.,
  Sastry, G., Askell, A., Mishkin, P., Clark, J., Krueger, G., and Sutskever,
  I.}
\newblock Learning transferable visual models from natural language
  supervision, 2021.

\bibitem{denseclip}
{\sc Rao, Y., Zhao, W., Chen, G., Tang, Y., Zhu, Z., Huang, G., Zhou, J., and
  Lu, J.}
\newblock Denseclip: Language-guided dense prediction with context-aware
  prompting.
\newblock In {\em CVPR\/} (2022).

\bibitem{livevqa}
{\sc Seshadrinathan, K., Soundararajan, R., Bovik, A.~C., and Cormack, L.~K.}
\newblock Study of subjective and objective quality assessment of video.
\newblock {\em IEEE TIP 19}, 6 (2010), 1427--1441.

\bibitem{vqc}
{\sc Sinno, Z., and Bovik, A.~C.}
\newblock Large-scale study of perceptual video quality.
\newblock {\em IEEE TIP 28}, 2 (2019), 612--627.

\bibitem{svqa}
{\sc Sun, W., Min, X., Lu, W., and Zhai, G.}
\newblock A deep learning based no-reference quality assessment model for ugc
  videos.
\newblock {\em arXiv preprint arXiv:2204.14047\/} (2022).

\bibitem{fastdvdnet}
{\sc Tassano, M., Delon, J., and Veit, T.}
\newblock Fastdvdnet: Towards real-time deep video denoising without flow
  estimation.
\newblock In {\em Proceedings of the IEEE/CVF Conference on Computer Vision and
  Pattern Recognition (CVPR)\/} (June 2020).

\bibitem{yfcc}
{\sc Thomee, B., Shamma, D.~A., Friedland, G., Elizalde, B., Ni, K., Poland,
  D., Borth, D., and Li, L.-J.}
\newblock Yfcc100m: The new data in multimedia research.
\newblock {\em Commun. ACM 59}, 2 (2016), 64–73.

\bibitem{videval}
{\sc Tu, Z., Wang, Y., Birkbeck, N., Adsumilli, B., and Bovik, A.~C.}
\newblock Ugc-vqa: Benchmarking blind video quality assessment for user
  generated content.
\newblock {\em IEEE TIP 30\/} (2021), 4449--4464.

\bibitem{rapique}
{\sc Tu, Z., Yu, X., Wang, Y., Birkbeck, N., Adsumilli, B., and Bovik, A.~C.}
\newblock Rapique: Rapid and accurate video quality prediction of user
  generated content.
\newblock {\em IEEE Open Journal of Signal Processing 2\/} (2021), 425--440.

\bibitem{matchhistogram}
{\sc Vonikakis, V., Subramanian, R., Arnfred, J., and Winkler, S.}
\newblock A probabilistic approach to people-centric photo selection and
  sequencing.
\newblock {\em IEEE Transactions on Multimedia 19}, 11 (2017), 2609--2624.

\bibitem{csiqvqa}
{\sc Vu, P.~V., and Chandler, D.~M.}
\newblock Vis3: an algorithm for video quality assessment via analysis of
  spatial and spatiotemporal slices.
\newblock {\em Journal of Electronic Imaging 23\/} (2014).

\bibitem{jpeg}
{\sc Wallace, G.~K.}
\newblock The jpeg still picture compression standard.
\newblock {\em Commun. ACM 34}, 4 (1991), 30–44.

\bibitem{icme2021}
{\sc Wang, H., Li, G., Liu, S., and Kuo, C.-C.~J.}
\newblock Icme 2021 ugc-vqa challenge.

\bibitem{clipiqa}
{\sc Wang, J., Chan, K. C.~K., and Loy, C.~C.}
\newblock Exploring clip for assessing the look and feel of images, 2022.

\bibitem{ytugccc}
{\sc Wang, Y., Inguva, S., and Adsumilli, B.}
\newblock Youtube ugc dataset for video compression research.
\newblock In {\em 2019 MMSP\/} (2019).

\bibitem{rfugc}
{\sc Wang, Y., Ke, J., Talebi, H., Yim, J.~G., Birkbeck, N., Adsumilli, B.,
  Milanfar, P., and Yang, F.}
\newblock Rich features for perceptual quality assessment of ugc videos.
\newblock In {\em CVPR\/} (June 2021), pp.~13435--13444.

\bibitem{simvlm}
{\sc Wang, Z., Yu, J., Yu, A.~W., Dai, Z., Tsvetkov, Y., and Cao, Y.}
\newblock Simvlm: Simple visual language model pretraining with weak
  supervision.
\newblock In {\em ICLR\/} (2022).

\bibitem{h264}
{\sc Wiegand, T.}
\newblock Draft itu-t recommendation and final draft international standard of
  joint video specification.

\bibitem{fastvqa}
{\sc Wu, H., Chen, C., Hou, J., Liao, L., Wang, A., Sun, W., Yan, Q., and Lin,
  W.}
\newblock Fast-vqa: Efficient end-to-end video quality assessment with fragment
  sampling.
\newblock In {\em ECCV\/} (2022).

\bibitem{fastervqa}
{\sc Wu, H., Chen, C., Liao, L., Hou, J., Sun, W., Yan, Q., Gu, J., and Lin,
  W.}
\newblock Neighbourhood representative sampling for efficient end-to-end video
  quality assessment.

\bibitem{discovqa}
{\sc Wu, H., Chen, C., Liao, L., Hou, J., Sun, W., Yan, Q., and Lin, W.}
\newblock Discovqa: Temporal distortion-content transformers for video quality
  assessment.

\bibitem{dover}
{\sc Wu, H., Liao, L., Chen, C., Hou, J., Wang, A., Sun, W., Yan, Q., and Lin,
  W.}
\newblock Disentangling aesthetic and technical effects for video quality
  assessment of user generated content.

\bibitem{buonavista}
{\sc Wu, H., Liao, L., Hou, J., Chen, C., Zhang, E., Wang, A., Sun, W., Yan,
  Q., and Lin, W.}
\newblock Exploring opinion-unaware video quality assessment with semantic
  affinity criterion.
\newblock In {\em ICME\/} (2023).

\bibitem{bvqiplus}
{\sc Wu, H., Liao, L., Wang, A., Chen, C., Hou, J.~H., Zhang, E., Sun, W.~S.,
  Yan, Q., and Lin, W.}
\newblock Towards robust text-prompted semantic criterion for in-the-wild video
  quality assessment, 2023.

\bibitem{internetvqa}
{\sc Xu, J., Li, J., Zhou, X., Zhou, W., Wang, B., and Chen, Z.}
\newblock Perceptual quality assessment of internet videos.
\newblock In {\em ACM MM\/} (2021).

\bibitem{piaadataset}
{\sc Yang, Y., Xu, L., Li, L., Qie, N., Li, Y., Zhang, P., and Guo, Y.}
\newblock Personalized image aesthetics assessment with rich attributes.
\newblock In {\em CVPR\/} (2022), pp.~19861--19869.

\bibitem{pvq}
{\sc Ying, Z., Mandal, M., Ghadiyaram, D., and Bovik, A.}
\newblock Patch-vq: 'patching up' the video quality problem.
\newblock In {\em CVPR\/} (2021).

\bibitem{paq2piq}
{\sc Ying, Z., Niu, H., Gupta, P., Mahajan, D., Ghadiyaram, D., and Bovik, A.}
\newblock From patches to pictures (paq-2-piq): Mapping the perceptual space of
  picture quality.
\newblock In {\em CVPR\/} (2020).

\bibitem{coca}
{\sc Yu, J., Wang, Z., Vasudevan, V., Yeung, L., Seyedhosseini, M., and Wu, Y.}
\newblock Coca: Contrastive captioners are image-text foundation models.

\bibitem{cadb}
{\sc Zhang, B., Niu, L., and Zhang, L.}
\newblock Image composition assessment with saliency-augmented multi-pattern
  pooling.
\newblock {\em arXiv preprint arXiv:2104.03133\/} (2021).

\bibitem{dbcnn}
{\sc Zhang, W., Ma, K., Yan, J., Deng, D., and Wang, Z.}
\newblock Blind image quality assessment using a deep bilinear convolutional
  neural network.
\newblock {\em IEEE TCSVT 30}, 1 (2020), 36--47.

\bibitem{liqe}
{\sc Zhang, W., Zhai, G., Wei, Y., Yang, X., and Ma, K.}
\newblock Blind image quality assessment via vision-language correspondence: A
  multitask learning perspective.
\newblock In {\em IEEE Conference on Computer Vision and Pattern Recognition\/}
  (2023).

\bibitem{clip3dqa}
{\sc Zhang, Z., Sun, W., Zhou, Y., Wu, H., Li, C., Min, X., and Liu, X.}
\newblock Advancing zero-shot digital human quality assessment through
  text-prompted evaluation, 2023.

\bibitem{mdvqa}
{\sc Zhang, Z., Wu, W., Sun, W., Tu, D., Lu, W., Min, X., Chen, Y., and Zhai,
  G.}
\newblock Md-vqa: Multi-dimensional quality assessment for ugc live videos.
\newblock In {\em Proceedings of the IEEE Conference on Computer Vision and
  Pattern Recognition (CVPR)\/} (2023).

\bibitem{coop}
{\sc Zhou, K., Yang, J., Loy, C.~C., and Liu, Z.}
\newblock Learning to prompt for vision-language models.
\newblock {\em International Journal of Computer Vision (IJCV)\/} (2022).

\bibitem{metaiqa}
{\sc Zhu, H., Li, L., Wu, J., Dong, W., and Shi, G.}
\newblock {MetaIQA:} deep meta-learning for no-reference image quality
  assessment.
\newblock In {\em Proceedings of the IEEE Conference on Computer Vision and
  Pattern Recognition (CVPR)\/} (Jun. 2020), pp.~14143--14152.

\end{thebibliography}
}

\end{document}